\documentclass[conference,onecolumn]{IEEEtran}
\usepackage[utf8]{inputenc}
\usepackage{amsmath,graphicx}
\usepackage{cite}
\usepackage{graphicx}
\usepackage{caption}
\usepackage{subcaption}
\usepackage{tikz}
\usepackage{subcaption}
\usetikzlibrary{positioning,arrows.meta}
\usepackage{tikz}
\usetikzlibrary{positioning}
\usepackage{booktabs}
\usetikzlibrary{shapes.geometric, arrows.meta, positioning}
\usepackage{multirow}
\usepackage{multicol}
\usepackage{url}  

\tikzstyle{block} = [rectangle, draw, minimum height=1.2cm, minimum width=2.8cm, align=center]
\tikzstyle{arrow} = [thick, ->, >=stealth]

\title{SpectraSentinel: LightWeight Dual‐Stream Real‐Time Drone Detection, Tracking and Payload Identification}

\author{
\IEEEauthorblockN{Shahriar Kabir\IEEEauthorrefmark{1}, Istiak Ahmmed Rifti\IEEEauthorrefmark{1}, H.M. Shadman Tabib\IEEEauthorrefmark{1},\\
Mushfiqur Rahman\IEEEauthorrefmark{1}, Sadatul Islam Sadi\IEEEauthorrefmark{1}, Hasnaen Adil\IEEEauthorrefmark{1}}\\
\IEEEauthorblockN{Ahmed Mahir Sultan Rumi\IEEEauthorrefmark{1},  Ch. Md. Rakin Haider\IEEEauthorrefmark{1}}\\
\IEEEauthorblockA{\IEEEauthorrefmark{1}Department of Computer Science and Engineering,\\
Bangladesh University of Engineering and Technology, Dhaka, Bangladesh\\
kabirshahriar468@gmail.com}
}

\begin{document}
\maketitle

\begin{abstract}
The proliferation of drones in civilian airspace has raised urgent security concerns, calling for robust real-time surveillance systems. We propose a dual-stream drone monitoring approach targeting the 2025 VIP Cup challenge tasks of drone detection, tracking, and payload identification. Our system employs separate You Only Look Once v11-nano (YOLOv11n) object detectors on parallel infrared (thermal) and visible (RGB) data streams without early fusion. This design allows each model to be optimized for its modality’s unique characteristics, addressing the challenges of small flying objects in diverse conditions. We tailor data preprocessing and augmentation strategies to each domain – e.g. restricting color jitter for IR imagery – and fine-tune training hyperparameters to improve detection of tiny drones and payloads under heavy noise, low light, and motion blur. The resulting lightweight YOLOv11n models achieve high detection accuracy for distinguishing drones from birds and classifying payload types, while operating within real-time constraints. In this report, we detail the motivation for a dual-modality solution, the specialized training pipeline for each sensor, and the architecture and optimizations that enable accurate, efficient drone surveillance using RGB and IR streams.
\end{abstract}

\section{Introduction}
Unmanned aerial vehicles (UAVs) have rapidly gained popularity for applications ranging from environmental monitoring to delivery. However, the misuse of drones poses significant public safety and privacy risks, including unauthorized aerial surveillance and malicious payload delivery~\cite{nature}. The ability to detect and track rogue drones – and identify any payloads they carry – in real time is now critical for security in sensitive areas. This is a challenging task: drones are often small, fast-moving objects in cluttered outdoor environments, leading to limited feature representation and background noise that complicate their detection. Moreover, computational constraints demand efficient algorithms deployable on edge devices or surveillance platforms.

Most vision-based drone detection systems rely on conventional RGB cameras, but visible-spectrum imaging can fail under adverse conditions such as nighttime, fog, or glare. In such scenarios, infrared (IR) thermal cameras offer complementary information by capturing the heat signatures of objects, enabling more robust detection when optical cameras are hindered. For example, an IR sensor can reveal a drone’s presence in darkness or haze that obscures it in RGB footage. On the other hand, IR imagery alone lacks the spatial detail and color texture of RGB images, which can be crucial for recognizing small payloads or distinguishing drones from birds. Thus, the two modalities have inherent complementary strengths.

Prior research has shown that combining visible and thermal data can improve detection accuracy and reliability in low-light conditions~\cite{arxivFusion}. Indeed, an ideal solution would fuse RGB and IR cues to exploit both thermal contrast and visual detail. However, designing an effective unified multispectral detector is non-trivial, requiring careful balancing of fusion strategies and modality contributions~\cite{arxivFusion}. In this work, we adopt a simpler dual-stream approach, processing IR and RGB inputs with separate models. This allows us to tailor the detection pipeline to each domain’s characteristics without the complexity of early fusion, while still enabling a late integration of results if needed.

Another motivation for multi-modal sensing in drone surveillance is payload identification. Drones can carry hazardous payloads (e.g. weapons, contraband, explosives), so recognizing the type of payload in real time is essential for threat assessment. RGB cameras provide visual cues about a payload’s shape or color, whereas IR can highlight heat-emitting components of a payload (such as powered devices or warm contents). A combination of both can improve payload detection, especially when payloads are small or camouflaged. The competition dataset reflects these needs, providing paired thermal and visual images of drones with either harmful or normal payloads. While sensor fusion would ideally enhance payload recognition, our system currently evaluates payload imagery from each modality separately to simplify the model design. We note that standalone IR or RGB analysis may miss certain payload cues (e.g. a non-heated but visually distinctive item, or vice versa), but separating streams allows independent optimization and analysis of each sensor’s performance.

Recent advances in computer vision and deep learning have started to tackle drone detection in challenging conditions. Traditional radar or RF-based detection methods exist, but vision-based methods are attractive for their high resolution and passive sensing. Modern object detection networks like You Only Look Once (YOLO) have demonstrated success in fast, accurate detection of drones and other small objects~\cite{yoloOrig}. However, many object detectors are developed on everyday RGB imagery and do not fully leverage the unique characteristics of thermal infrared images~\cite{mdpiThermal}. Thermal sensors produce grayscale heat maps with different noise profiles and contrast properties than color cameras, meaning detection algorithms must be adapted for IR data~\cite{mdpiThermal}.

Some works have proposed specialized lightweight YOLO variants for UAV detection. For instance, Wang et al. introduced a UAV-YOLOv8 model with multi-branch heads to improve small-object detection in drone footage~\cite{uavyolo}, and Zhao et al. developed G-YOLO with a modified YOLOv8 backbone and depthwise convolutions to better suit infrared UAV images~\cite{gyolo}. These efforts underscore the importance of tailoring architectures and training strategies to the domain (visible vs thermal) to achieve robust performance.

Building on this trend, we employ the latest YOLO family model – YOLOv11 – as the core of our detection system. We choose the ultra-compact YOLOv11n (nano) variant to meet real-time processing needs. Despite its small size, YOLOv11n provides strong baseline accuracy, as evidenced by recent studies using it for drone detection. Our approach consists of two parallel YOLOv11n detectors: one trained on IR images and one on RGB images. Each is trained to detect drones and distinguish them from confounding objects (e.g. birds) in its respective domain. Detections from these models can then be used jointly to cross-verify drone sightings or to hand off tracking from one spectrum to the other in difficult conditions.

We also train dedicated YOLOv11n models for payload detection, formulated as a two-class object detection problem (harmful vs. benign payload) on infrared and visible imagery respectively. By keeping the RGB and IR pipelines separate, we simplify the training process and exploit domain-specific data augmentations and hyperparameters for each. The remainder of this paper details our system design and training methodology. In the Approach section, we describe the data preprocessing steps, the YOLOv11n architecture and why it suits our application, the training strategies (including augmentation and optimization choices for IR vs RGB), and how we perform tracking and payload classification using the trained models.

\section{Approach}

\subsection{Data Preprocessing and Augmentation}

We trained and evaluated separate detection models for the IR and RGB datasets provided in the VIP Cup 2025 challenge. Each dataset contains annotated drone images (with drones and birds labeled) and a payload identification subset, with a consistent resolution of 320×256 pixels. Before training, we organized the data into the required YOLO format: images in each split (train/val/test) and corresponding annotation text files containing bounding boxes and class labels for each object.

Because YOLOv11n expects three-channel input, the single-channel IR images were replicated across 3 channels to mimic an RGB format (while preserving the thermal intensity values)~\cite{grayscale3ch}. No other color space conversions were needed, since IR images are essentially grayscale intensity maps. We maintained the native image resolution during training, padding or letterboxing as needed to fit the 320×256 frames into the network’s input dimension (320×320) without distorting aspect ratio.

Domain-specific data augmentation was a key design choice to handle the challenging conditions of each modality. The competition data already includes various distortions and adverse scenarios – such as noise, blur, low illumination, and camera instability – so we adopted a conservative augmentation strategy to avoid over-complicating training examples.

For the RGB model, we applied only minimal color jitter (hue shift of at most 0.5\%, saturation variation 10\%, and brightness variation 10\%) because the dataset contains many low-light and cloudy scenes where excessive color augmentation could be unrealistic. The IR model uses no hue or saturation augmentation at all, since thermal images lack color information~\cite{thermalAugment}. We only allow a modest variation in IR image brightness (value scale ±20\%) to account for different thermal contrasts.

Geometric augmentations were also kept limited for both domains: we avoided any rotation or shear transformations (the drone videos already include arbitrary camera angles and some instability) and used only small random translations (up to 5\% shift) and scaling (±10\%). Vertical flips were disabled (since the notion of “up” vs “down” may be fixed for aerial footage), while horizontal flips were applied with 30\% probability to double the range of viewing angles.

We found that aggressive augmentations (like large rotations or heavy distortions) were unnecessary given that the dataset itself provides challenging examples; instead, our strategy focused on mosaic and copy-paste augmentations to improve small object detection~\cite{yolov5augment}. We enabled MOSAIC augmentation with a high probability (80\%), which randomly combines four images during training. This helps expose the model to varied backgrounds and multi-object scenarios, addressing cases like drones appearing alongside birds or against complex terrain.

Additionally, we used a moderate amount of copy-paste augmentation (20\% chance), wherein small objects (drones or birds) are cut from one image and pasted into another. This specifically boosts the occurrence of tiny flying objects in varied contexts and improves the detector’s ability to recognize drones even when they appear in swarms or near other distractors.

Overall, our augmentation policy was tuned to each modality’s needs: the IR pipeline preserves thermal patterns (no hue/sat shifts) and avoids geometric warping, while the RGB pipeline applies slight color variability. Both pipelines rely on mosaic and copy-paste to enrich the training data without introducing artificial artifacts that could confuse the models.

Prior to feeding images into the network, we normalize pixel values and, in the case of the payload dataset, ensure that images without any payload still have corresponding “no object” labels to not confuse the model. The payload identification data was handled similarly in separate IR and RGB streams. Since payloads are often much smaller than drones, we paid special attention during annotation parsing to include those small objects (some payloads are only a few pixels in size) and used the same augmentation strategy (minus color for IR) to slightly perturb payload appearances.

\section{Model Training and Inference Pipeline}

Our drone detection and payload identification system utilizes the YOLOv11n detector— the lightweight, nano variant of the YOLOv11 one-stage object detection architecture. YOLOv11 is known for its efficient CNN backbone, enhanced multi-scale detection head, and optimized attention mechanisms, providing strong performance for small-object detection in cluttered environments. Despite being compact, YOLOv11n achieves real-time inference speeds with high accuracy, making it ideal for deployment on edge devices and modest hardware.

\subsection{\textbf{Training Drone Bird}}

We trained separate YOLOv11n models for RGB and Infrared (IR) image modalities. Both models were trained independently using modality-specific training datasets to learn distinct visual and thermal signatures of drones and payloads. Payload identification models were separately trained to recognize payload types—harmful or normal—based on clearly annotated bounding boxes around payload regions.

The RGB and IR models utilized PyTorch’s Automatic Mixed Precision (AMP) to accelerate training, maintaining accuracy while improving efficiency. We standardized training input sizes to 320×320 to optimize computational resources and minimize unnecessary upscaling.

\begin{figure}[h]
    \centering
    \includegraphics[width=0.75\linewidth]{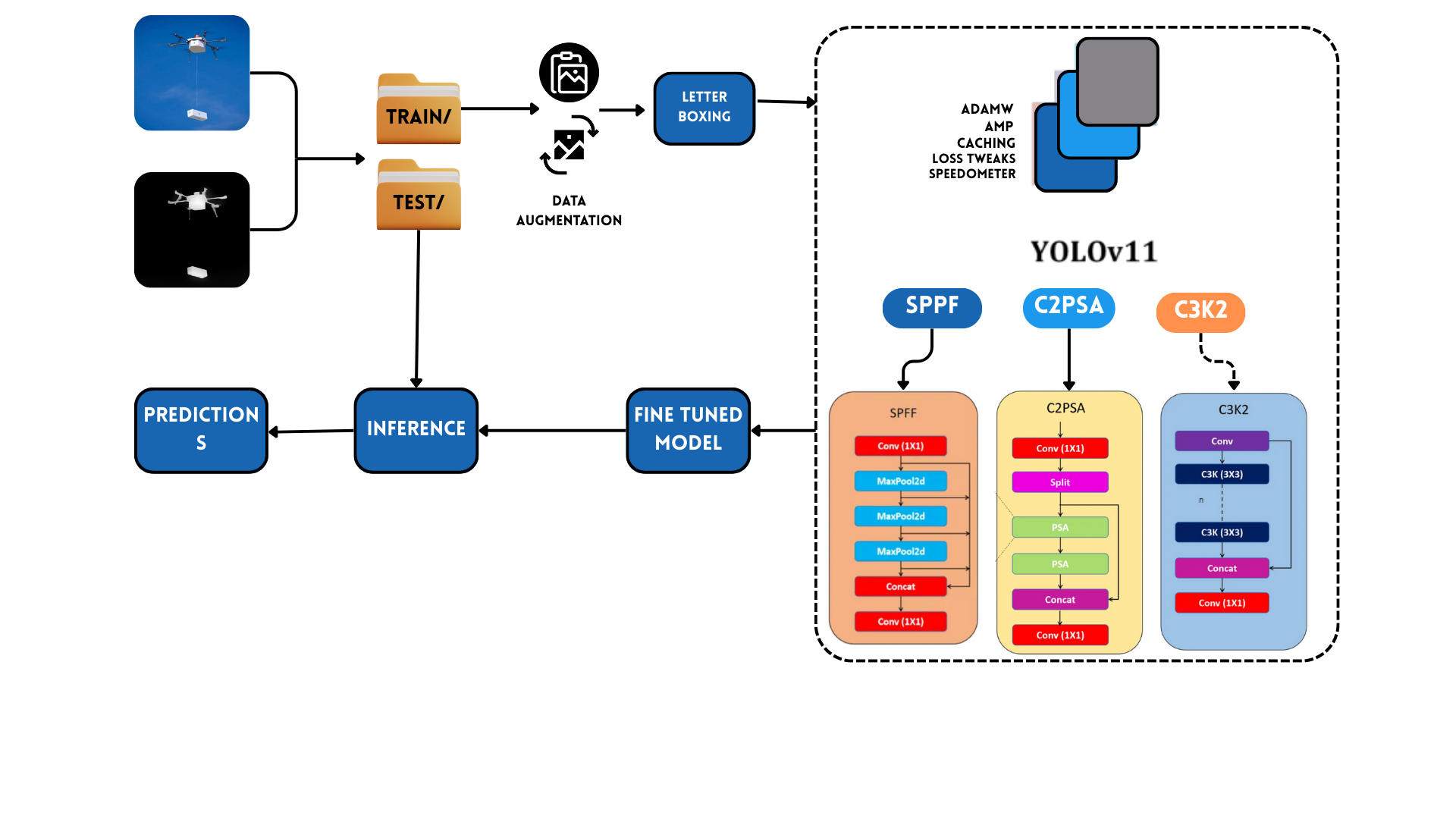}
    \caption{System overview with training and inference pipeline.}
    \label{fig:single_flow}
\end{figure}

During inference, our system dynamically handles three scenarios: RGB-only, IR-only, or combined RGB-IR inputs. The inference pipeline is as follows:

\begin{itemize}
\item \textbf{Both RGB and IR images available:} Each modality passes through its respective YOLOv11n backbone, producing separate detection outputs. The RGB detection is processed through Non-Maximum Suppression (NMS) to filter overlapping bounding boxes effectively, while IR detection leverages confidence-based activation, optimized for thermal signatures. Both outputs enter the Decision Layer independently, enabling complementary detection results from both modalities.
\item \textbf{Only IR image available:}  
The IR input is processed normally through the YOLO backbone. Simultaneously, a white image (RGB: 255,255,255) is passed as a placeholder to the RGB channel to maintain the integrity of the dual-input system architecture. The IR output proceeds to the Decision Layer via confidence-based detection, as NMS-based scoring is suboptimal in this scenario.

\item \textbf{Only RGB image available:}  
Similarly, the RGB input undergoes standard processing, while the IR backbone receives a white placeholder image. Detection results are subsequently refined using NMS activation, ensuring accurate bounding box selection before entering the Decision Layer.
\end{itemize}

The Decision Layer finalizes outputs from each channel based on their respective activation methods, producing bounding box detections and confidence scores. This approach efficiently adapts to single or dual-modality inputs without requiring complex feature-level fusion, significantly reducing computational overhead and latency.

\begin{figure}[h]
    \centering
    \begin{subfigure}[b]{0.9\linewidth}
        \includegraphics[width=\linewidth]{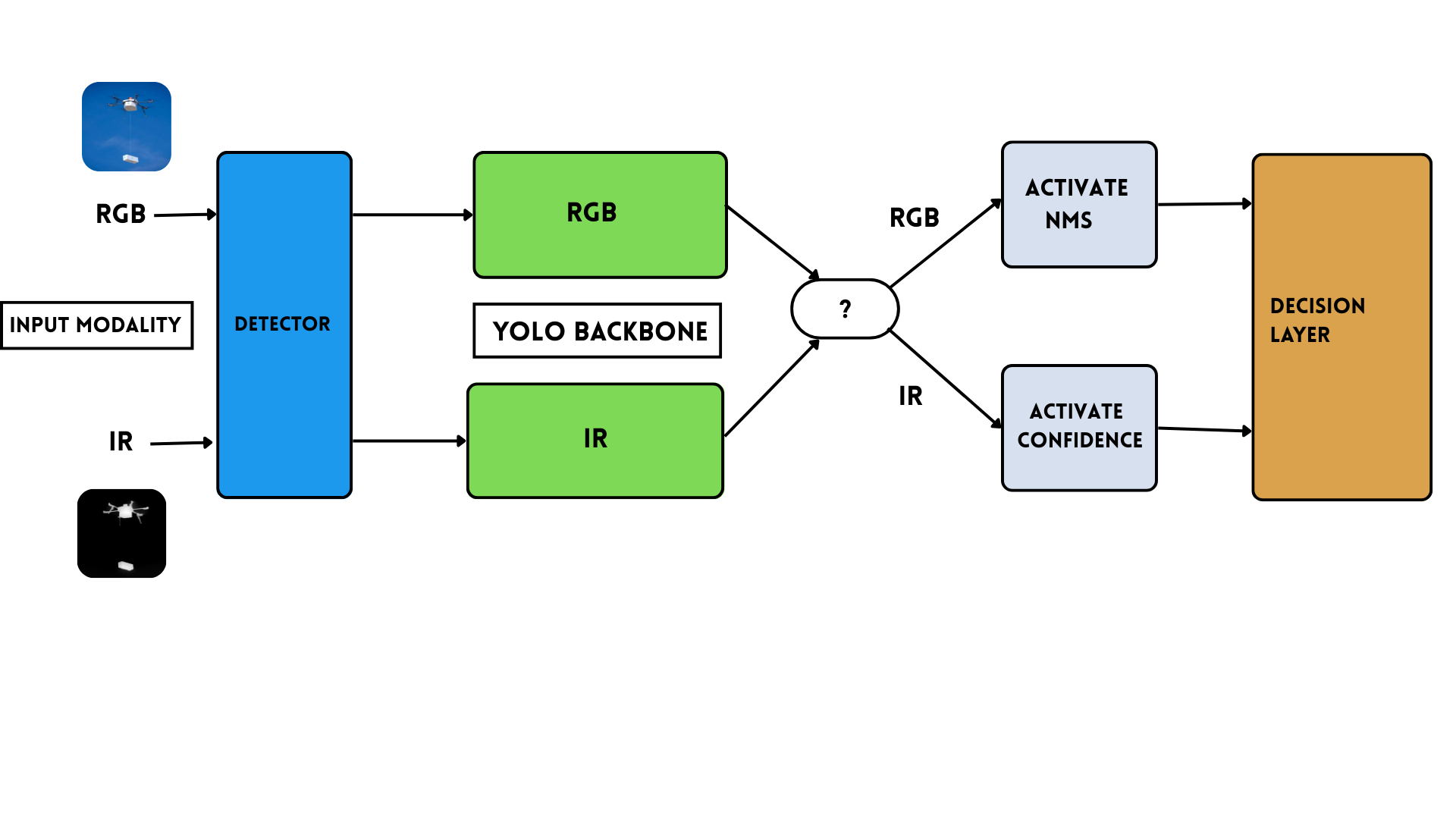}
        \caption{RGB-IR channel Inference}
        \label{fig:combined}
    \end{subfigure}

    \vspace{1em}

    \begin{subfigure}[b]{0.45\linewidth}
        \includegraphics[width=\linewidth]{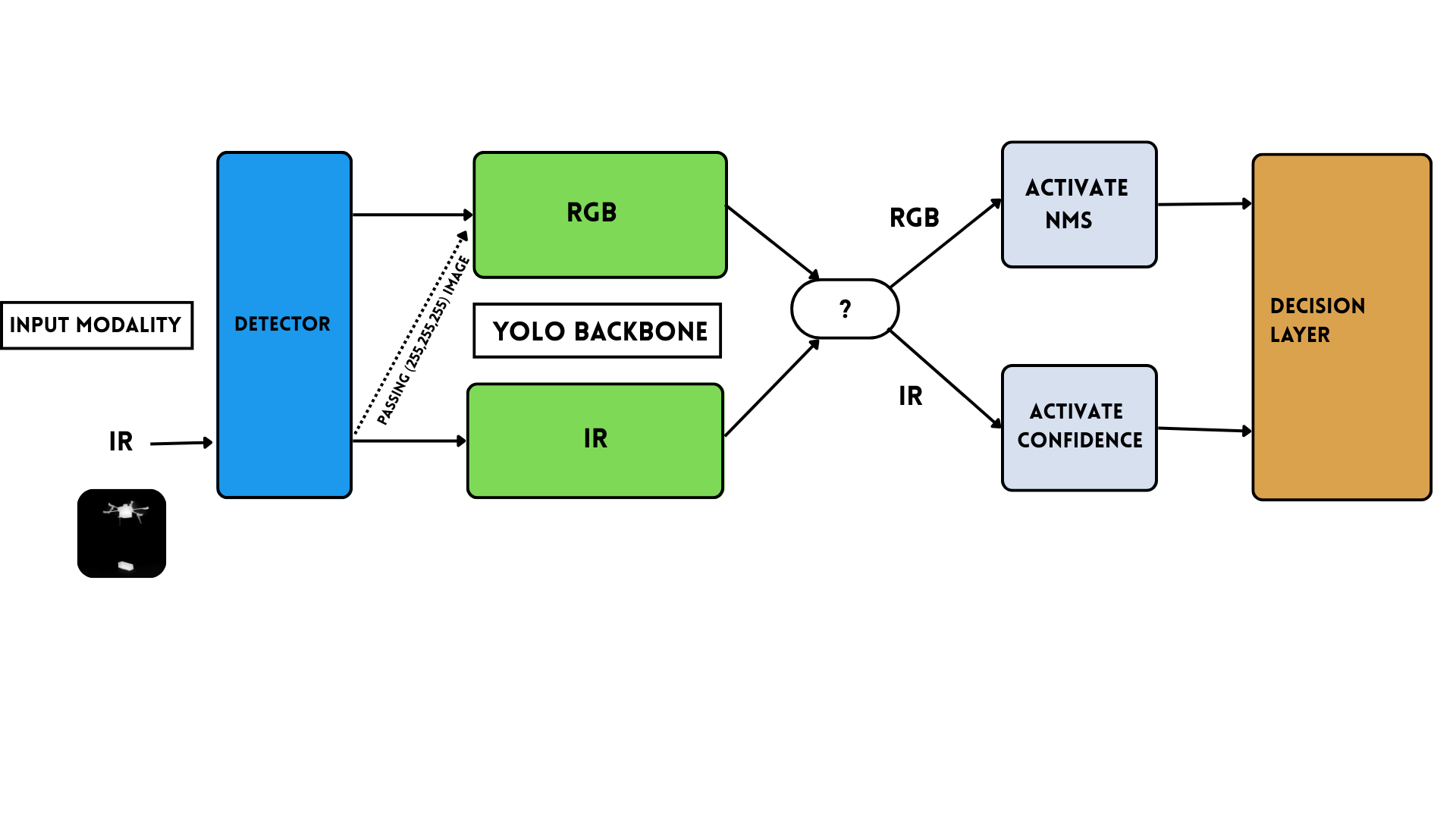}
        \caption{RGB-only inference}
        \label{fig:rgb_only}
    \end{subfigure}
    \hfill
    \begin{subfigure}[b]{0.45\linewidth}
        \includegraphics[width=\linewidth]{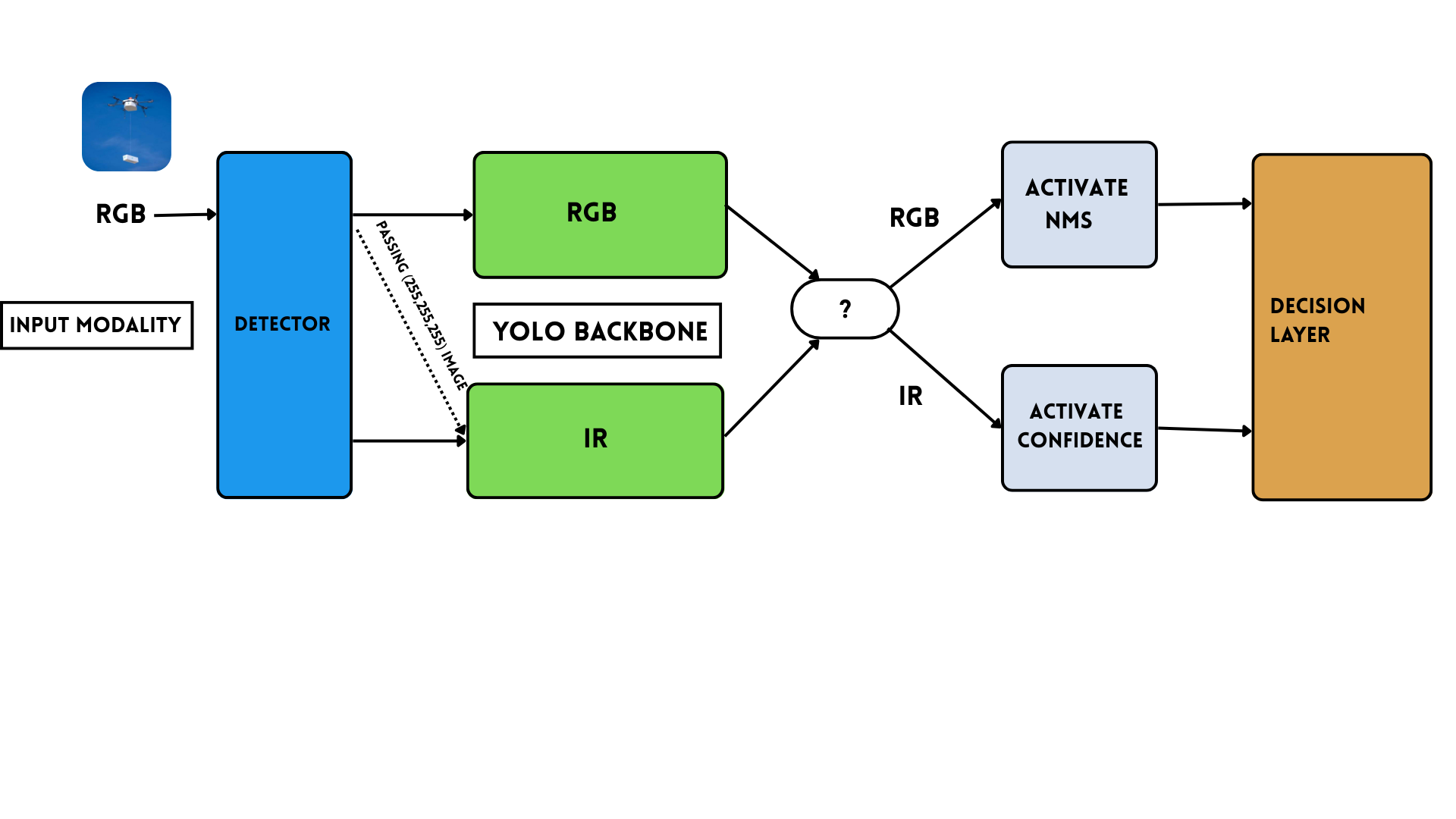}
        \caption{IR-only inference}
        \label{fig:ir_only}
    \end{subfigure}

    \caption{Inference modes: IR-only, RGB-only, and dual modality handled uniformly via placeholder injection and parallel backbone execution.If only IR is available , white image is passed to RGB channel (b). If only RGB is available , white image is passed to IR channel (c) .}
    \label{fig:panel_flow}
\end{figure}
\subsubsection*{\textbf{Real-Time Performance}}

Our implementation achieves real-time performance even on modest hardware. On NVIDIA RTX GPUs, the YOLOv11n models comfortably exceed 30 FPS for 320×256 video streams, processing single frames within milliseconds. CPU inference remains viable, achieving near-real-time performance beneficial for embedded systems or edge scenarios. By parallelizing RGB and IR backbone processing, our pipeline ensures balanced computational load and minimal latency.

\subsection{\textbf{Tracking and Payload Identification}}

Drone tracking is achieved using a lightweight Intersection-over-Union (IoU)-based tracker. After YOLOv11n detects drones in either the RGB or IR image stream, consecutive detections are linked by comparing bounding boxes from frame to frame. Tracks are maintained through short detection gaps (up to 10--15 frames) to handle brief occlusions or intermittent detection failures, ensuring reliable trajectory continuity.

For \textbf{payload identification}, we implement a modality-aware approach tailored for the availability of input streams:

\begin{itemize}
\item \textbf{When both RGB and IR modalities are available:} \
RGB and IR images each feed their respective YOLOv11n backbones, and their outputs jointly undergo Non-Maximum Suppression (NMS) before entering the final decision layer. This fusion strategy maximizes accuracy by combining complementary information from both modalities.
\item \textbf{When only the RGB modality is available:} \\ 
A grayscale version of the RGB input image is generated and passed to the IR backbone as a surrogate, ensuring both backbones receive valid inputs. Both outputs still proceed jointly to NMS and the decision layer, preserving consistency in inference regardless of IR input availability.
\begin{figure}[h]
    \centering
    \begin{subfigure}[b]{0.9\linewidth}
        \includegraphics[width=\linewidth]{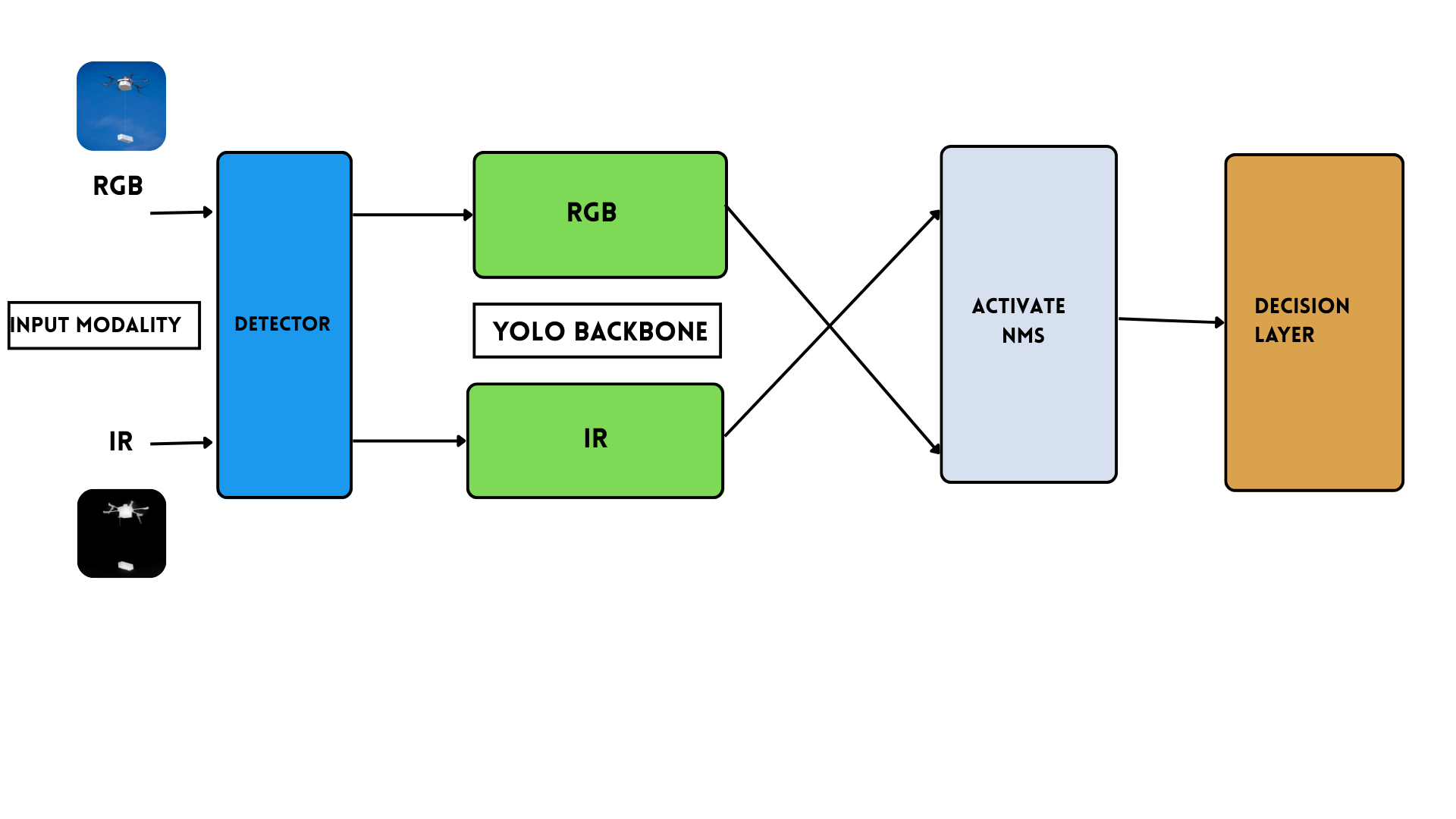}
        \caption{RGB-IR channel Inference}
        \label{fig:combined}
    \end{subfigure}

    \vspace{1em}

    \begin{subfigure}[b]{0.45\linewidth}
        \includegraphics[width=\linewidth]{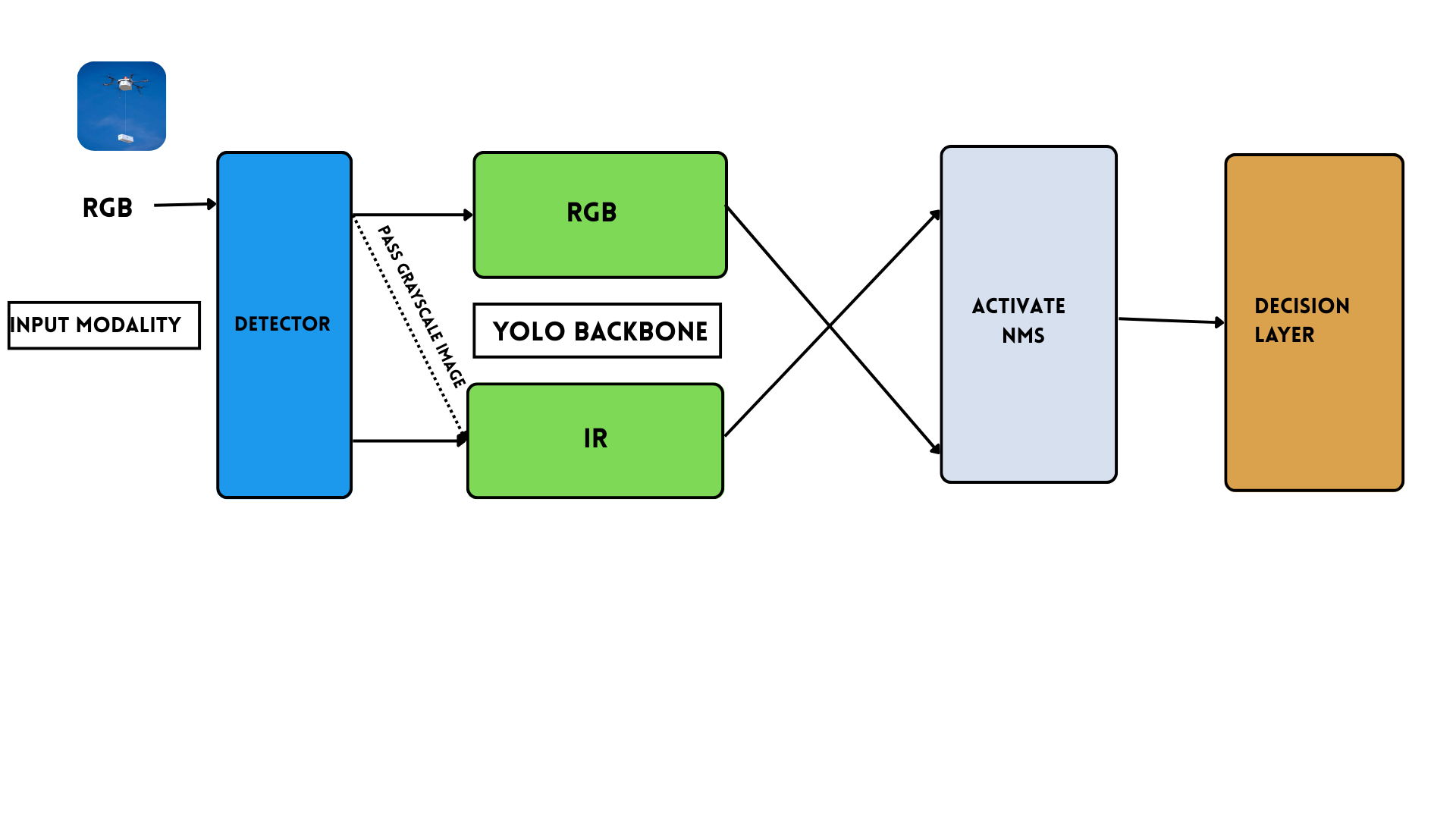}
        \caption{RGB-only inference}
        \label{fig:rgb_only}
    \end{subfigure}
    \hfill
    \begin{subfigure}[b]{0.45\linewidth}
        \includegraphics[width=\linewidth]{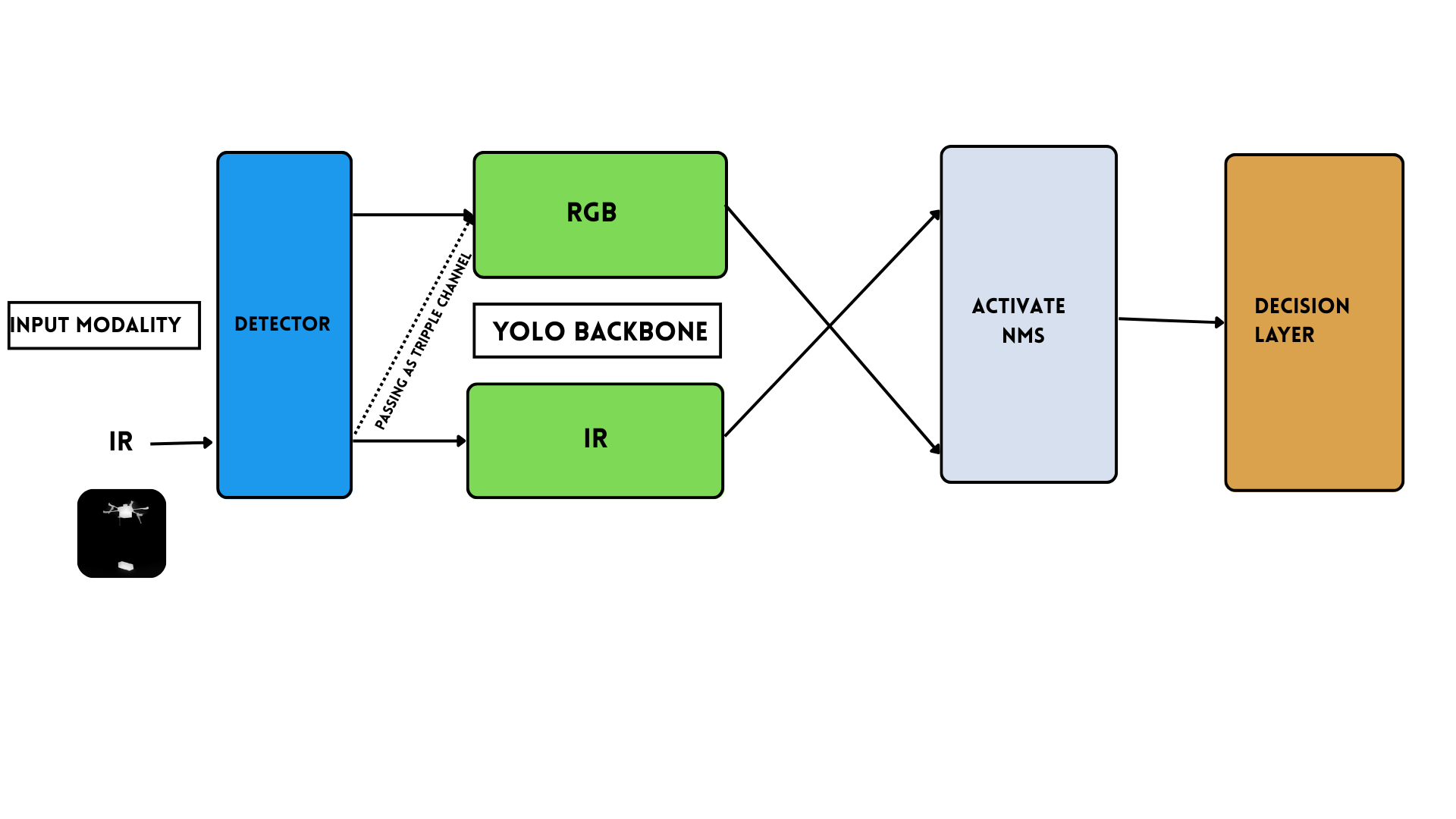}
        \caption{IR-only inference}
        \label{fig:ir_only}
    \end{subfigure}

    \caption{Inference modes in Payload Detection: IR-only, RGB-only, and dual modality handled uniformly via placeholder injection and parallel backbone execution.If only IR is available , triplet  image is passed to RGB channel (b). If only RGB is available , grayscale image is passed to IR channel (c) .}
    \label{fig:panel_flow}
\end{figure}

\item \textbf{When only the IR modality is available:} \\ 
The IR image is replicated across three channels (forming a pseudo-RGB triplet) and passed through the RGB backbone, while the IR backbone receives the original IR input. As before, the outputs undergo joint NMS processing, maintaining the system's structural uniformity.
\end{itemize}
\subsubsection*{\textbf{Real-Time Performance}}
Each backbone outputs bounding boxes and confidence scores independently, and subsequent payload classification models trained specifically for RGB and IR streams detect and categorize payloads (harmful or normal). Due to intrinsic modality differences, dual inputs provide complementary advantages---for example, IR excels in low-light or visually obscured scenarios, while RGB better captures visible detail and color information.

During practical inference, the fusion logic for payload identification employs a logical OR: if either RGB or IR payload detection is confident in classifying a payload as harmful, the payload is flagged accordingly. This approach effectively mitigates individual modality blind spots and significantly enhances payload classification reliability.

This structured yet flexible pipeline achieves robust drone surveillance performance by intelligently adapting to modality availability, effectively balancing accuracy, computational efficiency, and inference consistency.

\begin{figure}[htbp]
  \centering
  \begin{subfigure}[b]{\linewidth}
    \centering
    \resizebox{0.9\linewidth}{!}{%
    \begin{tikzpicture}[
        block/.style={rectangle, draw, minimum height=0.8cm, minimum width=2cm, align=center, font=\footnotesize},
        arrow/.style={thick,->,>=stealth},
        node distance=0.6cm and 0.4cm
      ]
      \node[block] (input)   {Video Input};
      \node[block] (extract) [right=of input]   {Frame Ext.};
      \node[block] (detect)  [right=of extract] {Detection};
      \node[block] (format)  [right=of detect]  {Formatting};
      \node[block] (track)   [right=of format]  {Tracking};
      \node[block] (dirmod)  [right=of track]   {Direction Est.};
      \node[block] (output)  [right=of dirmod]  {Annot. \& CSV};
      
      \node[block] (savef)   [below=1.2cm of output, xshift=-1cm] {Frame Save};
      \node[block] (savev)   [below=1.2cm of output, xshift=1cm] {Video Save};
      
      \draw[arrow] (input)   -- (extract);
      \draw[arrow] (extract) -- (detect);
      \draw[arrow] (detect)  -- (format);
      \draw[arrow] (format)  -- (track);
      \draw[arrow] (track)   -- (dirmod);
      \draw[arrow] (dirmod)  -- (output);
      
      \draw[arrow] (output) -- ++(0,-0.6cm) -| (savef);
      \draw[arrow] (output) -- ++(0,-0.6cm) -| (savev);
    \end{tikzpicture}
    }
    \caption{Core Video Tracking Pipeline}
    \label{fig:core_pipeline}
  \end{subfigure}
  
  \vspace{1.5em}
  
  \begin{subfigure}[b]{\linewidth}
    \centering
    \resizebox{0.95\linewidth}{!}{%
    \begin{tikzpicture}[
        block/.style={rectangle, draw, minimum height=1.2cm, minimum width=2.5cm, align=center, font=\footnotesize},
        cue/.style={rectangle, draw, minimum height=0.8cm, minimum width=2cm, align=center, font=\footnotesize},
        arrow/.style={thick,->,>=stealth},
        node distance=1.5cm
      ]
      \node[block] (hist) at (0,0) {Frame Buffer \\ \& History};
      
      \node[cue] (area)     at (4.5, 2)   {Area Trend};
      \node[cue] (centroid) at (4.5, 0.5) {Centroid Vel.};
      \node[cue] (scale)    at (4.5, -1)  {Scale Var.};
      \node[cue] (flow)     at (4.5, -2.5) {Sparse Flow};
      
      \node[block] (fusion) at (9, 0) {Fusion \&\\Temporal Smoothing};
      
      \draw[arrow] (hist.east) -- ++(0.5cm,0) |- (area.west);
      \draw[arrow] (hist.east) -- ++(0.8cm,0) |- (centroid.west);
      \draw[arrow] (hist.east) -- ++(1.1cm,0) |- (scale.west);
      \draw[arrow] (hist.east) -- ++(1.4cm,0) |- (flow.west);
      
      \draw[arrow] (area.east) -- ++(0.5cm,0) |- (fusion.west);
      \draw[arrow] (centroid.east) -- (fusion.west);
      \draw[arrow] (scale.east) -- ++(0.5cm,0) |- (fusion.west);
      \draw[arrow] (flow.east) -- ++(0.5cm,0) |- (fusion.west);
      
      \draw[arrow] (area.south) -- (centroid.north);
    \end{tikzpicture}
    }
    \caption{Multi-Cue Direction Estimation Module}
    \label{fig:cue_pipeline}
  \end{subfigure}
  
  \caption{Overall Video Tracking and Direction Estimation Architecture}
  \label{fig:overall_pipeline}
\end{figure}
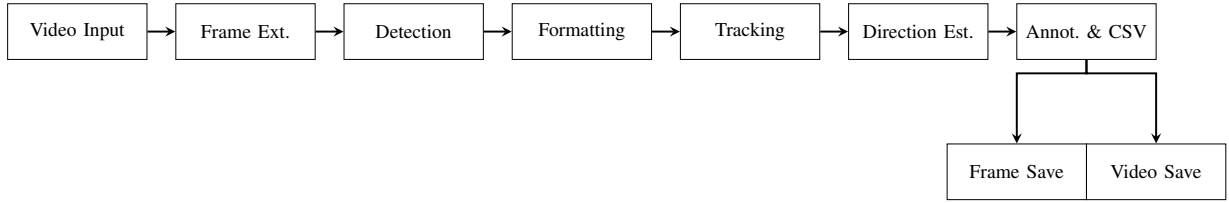
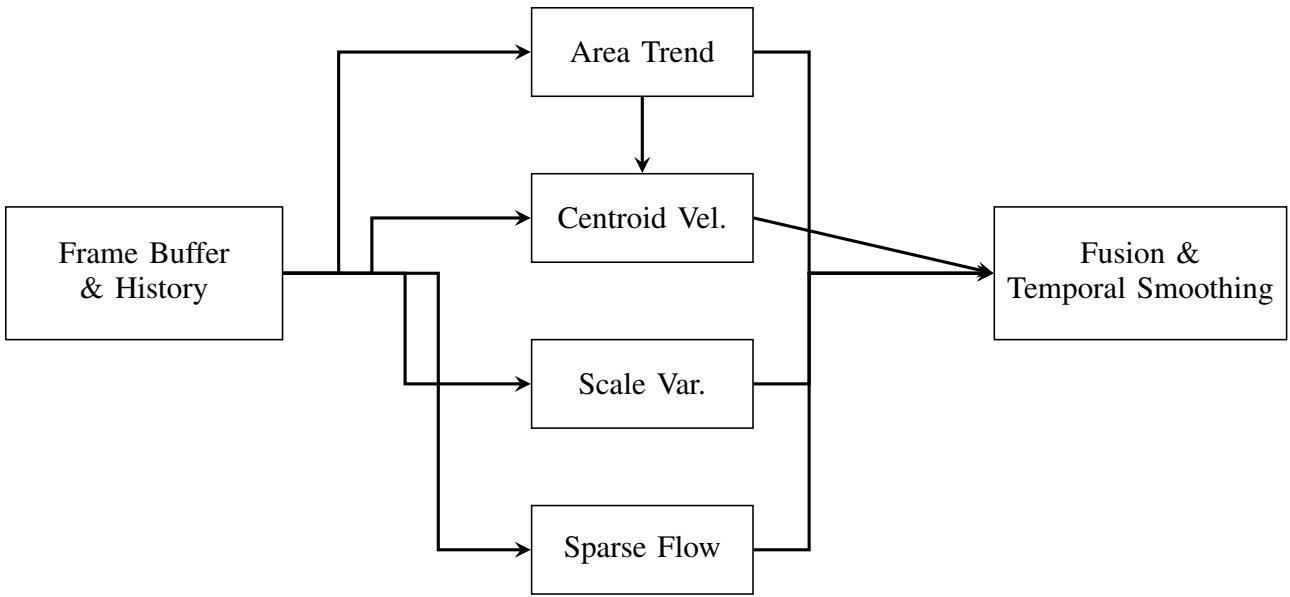

\subsection{\textbf{Video Tracking and Direction}}

\noindent
We propose a real-time, multi-stage pipeline for video-based drone tracking and direction estimation. This framework integrates fast object detection, appearance-based multi-object tracking, and a hybrid visual reasoning module to robustly infer the motion direction of aerial targets. Each component is designed for responsiveness and computational efficiency, with particular attention to how it scales in real-world deployment.

\begin{itemize}
    \item \textbf{Object Detection:} Each video frame is processed using a real-time object detector to localize aerial entities such as drones or birds. The detector produces class-specific bounding boxes with high speed and precision, enabling downstream modules to operate reliably under real-time constraints.

    \item \textbf{Multi-Object Tracking:} A tracker associates detected objects across frames using both motion prediction and appearance embeddings. This allows for stable identity maintenance over time, even in the presence of occlusions, viewpoint changes, or sudden object motion.

    \item \textbf{Direction Estimation:} Each tracked object is passed through a multi-cue direction estimator, which combines several visual indicators—including changes in object area, movement of object centroids, scale variation, and optical flow vectors. To maintain computational efficiency, the optical flow is applied selectively and sparsely over a small grid within the object region. This minimizes the per-frame workload, ensuring real-time operability even when multiple objects are tracked simultaneously.

    \item \textbf{Output Generation:} The final system output includes annotated video frames and structured logs capturing object tracks, spatial metadata, direction labels, and confidence levels. These outputs facilitate both visual inspection and downstream analytics such as trajectory forecasting or threat detection.
\end{itemize}

\noindent
\textbf{Performance Characteristics:}

The direction estimator is built with real-time performance as a central design goal. Among the most computationally intensive components is the optical flow computation, which has been carefully constrained by limiting the number of analyzed points and keeping the spatial search region small. This ensures that the motion estimation step remains efficient even in multi-object scenarios. Importantly, the estimator avoids redundant processing by handling each tracked object independently and reinitializing motion cues relative to the object’s updated position.

Memory usage is optimized by retaining only a short, fixed-length history of relevant features for each tracked object. Efficient queue-based data structures are employed to manage these histories, providing constant-time updates and avoiding memory bloat. This ensures that system performance remains stable across prolonged video sequences.

The estimator also employs lightweight arithmetic operations to evaluate visual trends, such as area changes or velocity. These operations are bounded and applied over short time windows, ensuring consistent and predictable frame-wise runtime regardless of video length. Logic for combining directional cues is similarly efficient, relying on simple statistical comparisons and minimal branching, without introducing significant computational overhead.

The algorithm's complexity grows linearly with the number of tracked objects per frame. However, because the work per object is capped and does not grow with video length, the system gracefully scales to increased target counts without compromising responsiveness. When object counts become large, potential enhancements such as batch processing of motion cues or GPU acceleration may be considered—but the current implementation performs robustly under typical aerial monitoring conditions.

Additional trade-offs are introduced to balance accuracy and speed. For example, very small objects, which are likely to be distant or noisy detections, are excluded from motion analysis to save processing time. Similarly, by using a uniform grid for motion tracking rather than dynamically selected feature points, the estimator reduces variability and avoids extra computation. Temporal smoothing and confidence aggregation further help stabilize outputs, reducing the impact of transient anomalies without adding significant delay or memory overhead.

\noindent
Overall, the system achieves real-time performance through careful design choices that minimize per-frame cost, bound memory use, and scale predictably with scene complexity. These characteristics make it well-suited for embedded or field deployment scenarios where computational resources are limited but fast, reliable tracking and directional inference are essential.

\section{Experiments and Results}
\subsection{Drone and Bird Detection}
In this study, we evaluated several dual-stream object detection architectures designed for drone and bird detection tasks under the VIP Cup 2025 Task 3 challenge, utilizing both RGB and infrared (IR) imaging modalities. Our primary objective was to investigate the effectiveness of different fusion approaches—namely mid-level fusion, transformer-based mid-level fusion, mid-to-late fusion, and late fusion—in improving detection performance within challenging multimodal surveillance scenarios.

\paragraph{Baseline and Architectural Variants.}
We employed and adapted the Ultralytics YOLO11 object detection framework to support dual-path processing for RGB and IR modalities. Our base architecture utilized dual independent YOLO11 backbones to extract multiscale features from each modality. We implemented several variants based on different feature fusion stages:

\begin{itemize}
\item \textbf{Mid-Fusion (Basic):} Features from both modalities at levels P3, P4, and P5 were concatenated and subsequently processed through SPPF and C2PSA modules before reaching the detection heads. This approach enabled efficient cross-modal spatial context integration while preserving individual modality features.
\item \textbf{Mid-to-Late Fusion:} Each modality underwent independent head processing up to the final layers, at which point feature maps at levels P3, P4, and P5 were concatenated just before the detection head. This method aimed at maintaining modality-specific representations longer.

\item \textbf{Mid-Fusion with Transformer Blocks:} Inspired by ICAFusion~\cite{ICAFusion}, standard concatenation was replaced with Transformer-based fusion blocks (\texttt{TransformerFusionBlock}) at levels P3, P4, and P5. This attention-based approach facilitated robust alignment of features across modalities.
\end{itemize}

\paragraph{Training Setup.}
All models were trained from scratch on the VIP Cup 2025 dataset, consisting of synchronized RGB and IR frames annotated with drone and bird labels. Models were trained for between 100 and 150 epochs, using a batch size of 16 and employing standard data augmentation techniques with cosine learning rate scheduling. Performance evaluation adhered to the standard COCO metric of mean Average Precision (mAP@0.5:0.95).

\begin{figure}[htbp]
    \centering
    \fbox{%
      \begin{minipage}{\dimexpr\linewidth-2\fboxsep-2\fboxrule\relax}
        \centering
        
        \begin{subfigure}[b]{0.32\linewidth}
            \centering
            \includegraphics[width=\linewidth]{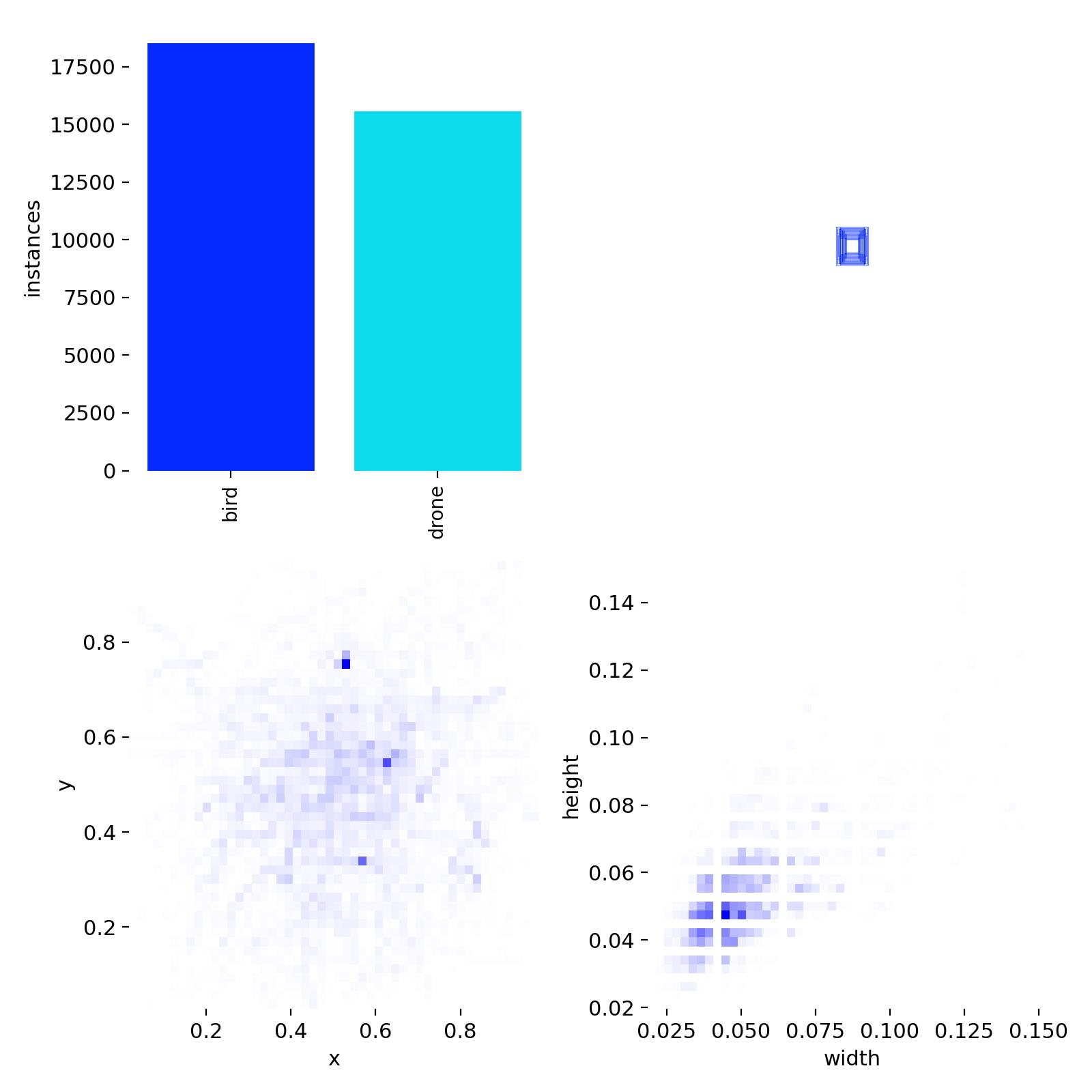}
            \caption{Class distribution \& bbox stats}
            \label{fig:panel:labels}
        \end{subfigure}%
        \hfill%
        \begin{subfigure}[b]{0.32\linewidth}
            \centering
            \includegraphics[width=\linewidth]{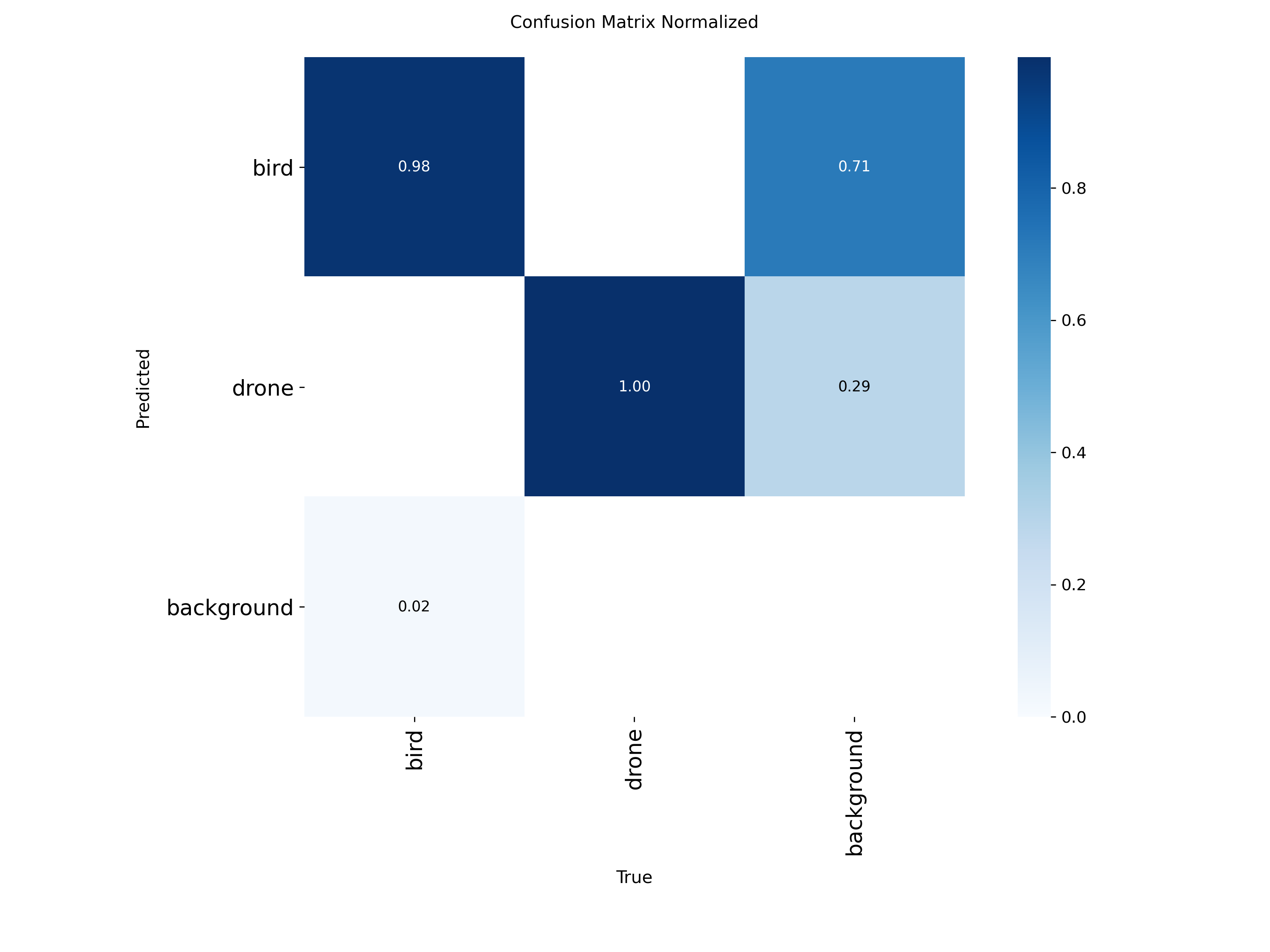}
            \caption{Normalized confusion matrix}
            \label{fig:panel:confusion}
        \end{subfigure}%
        \hfill%
        \begin{subfigure}[b]{0.32\linewidth}
            \centering
            \includegraphics[width=\linewidth]{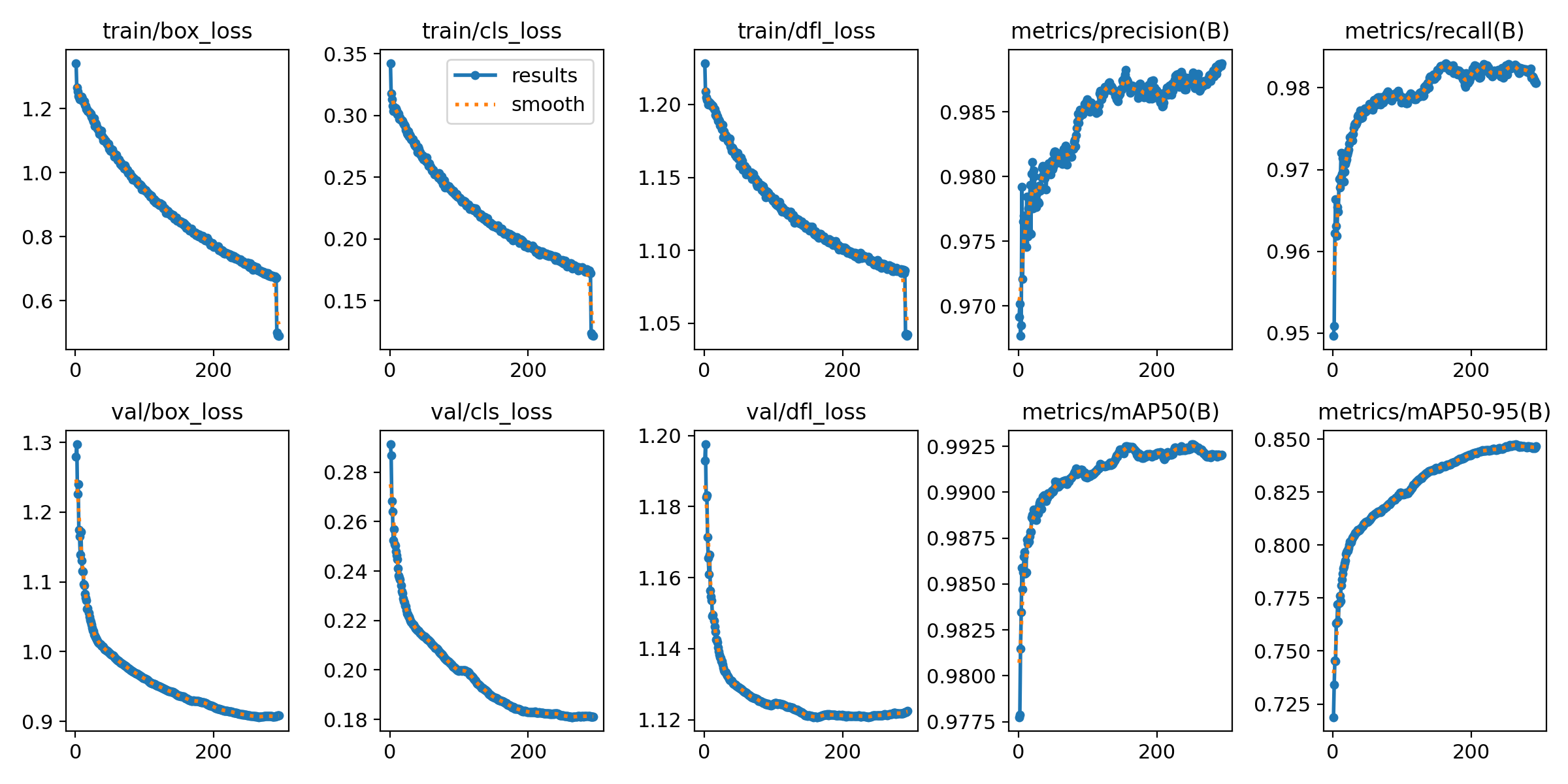}
            \caption{Training \& validation metrics}
            \label{fig:panel:results}
        \end{subfigure}
        
        \vspace{1em}
        
        \begin{subfigure}[b]{0.48\linewidth}
            \centering
            \includegraphics[width=\linewidth]{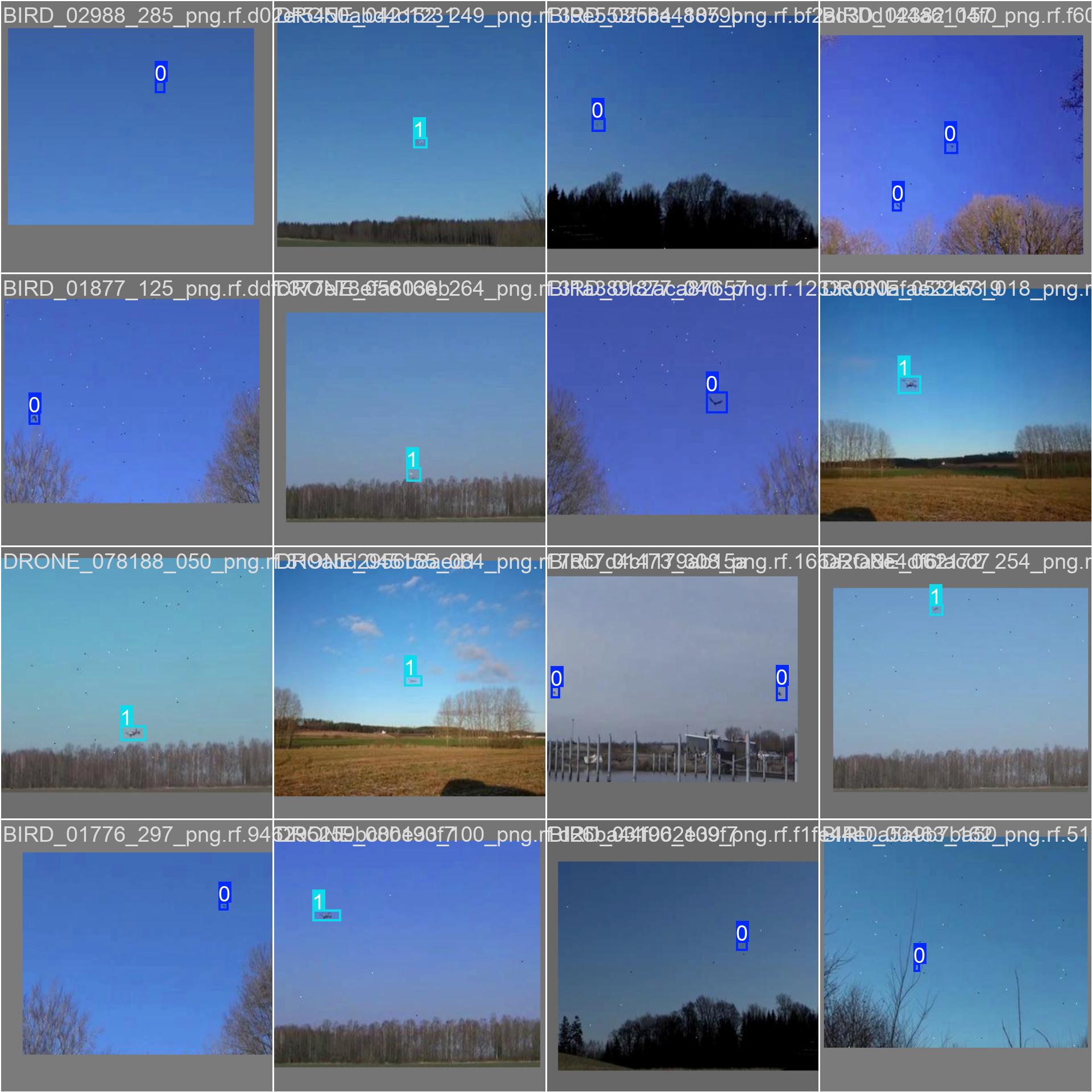}
            \caption{Example training batch}
            \label{fig:panel:train_batch}
        \end{subfigure}%
        \hfill%
        \begin{subfigure}[b]{0.48\linewidth}
            \centering
            \includegraphics[width=\linewidth]{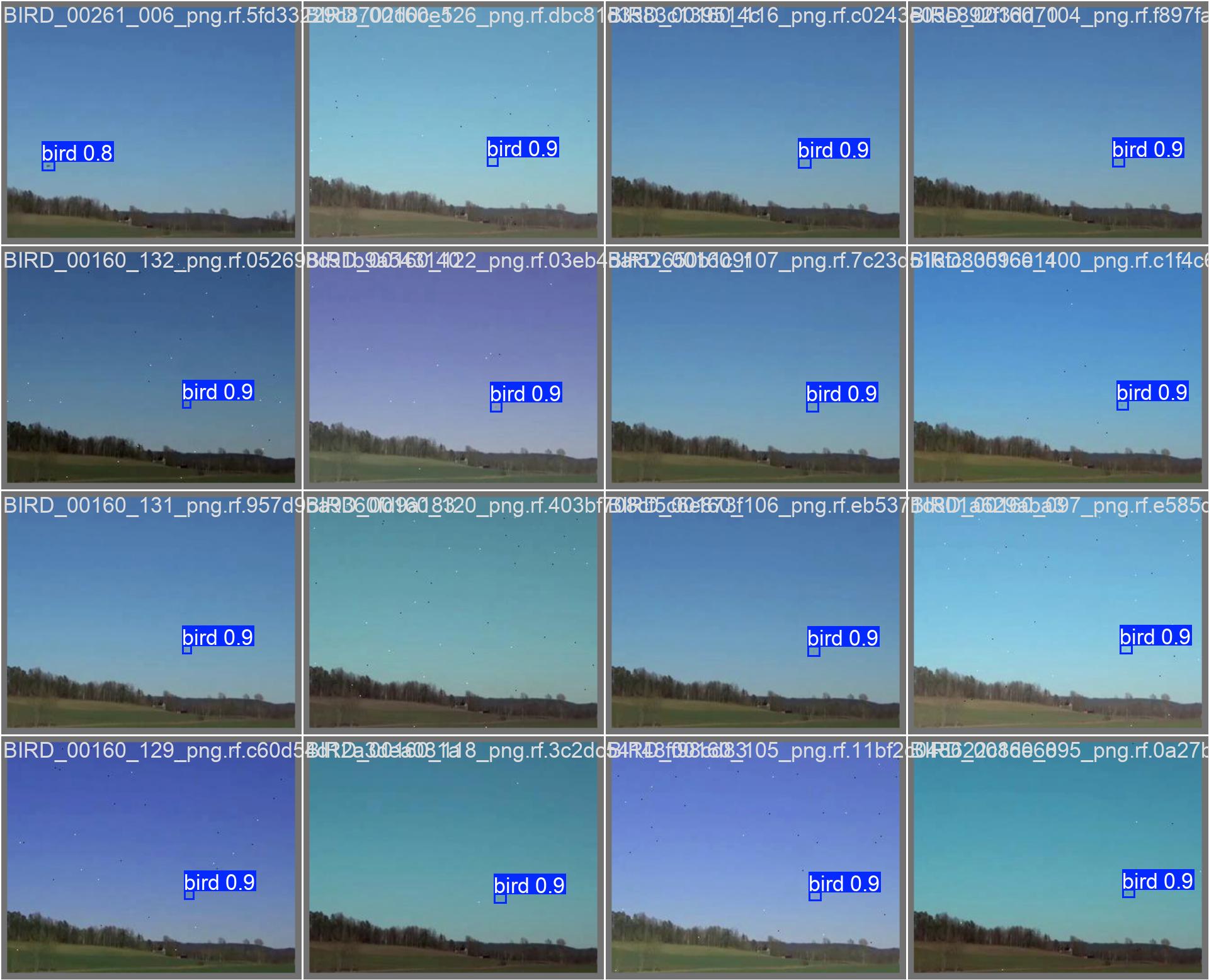}
            \caption{Validation predictions}
            \label{fig:panel:val_pred}
        \end{subfigure}
        
      \end{minipage}%
    }
    \caption[Dataset statistics and model performance overview for training RGB Drone Image]{%
      \textbf{Dataset Statistics and Model Performance Overview for training RGB Drone Image} 
      (a) Class distribution and bounding-box dimension statistics. 
      (b) Normalized confusion matrix. 
      (c) Evolution of training and validation losses \& metrics. 
      (d) Representative training batch with ground truth. 
      (e) Sample validation detections with confidence scores.%
    }
    \label{fig:dataset_and_results}
\end{figure}


\begin{table}[htbp]
  \centering
  \caption{Final Training \& Validation Results on RGB Dataset (Epoch 293)}
  \label{tab:rgb_results}
  \begin{tabular}{lccc ccc cccc}
    \toprule
    \multirow{2}{*}{\textbf{Epoch}} 
      & \multicolumn{3}{c}{\textbf{Training Losses}} 
      & \multicolumn{3}{c}{\textbf{Validation Losses}} 
      & \multicolumn{4}{c}{\textbf{Validation Metrics}} \\
    \cmidrule(r){2-4} \cmidrule(r){5-7} \cmidrule(r){8-11}
      & Box   & Cls   & DFL   
      & Box   & Cls   & DFL   
      & Prec. & Recall & mAP@0.5 & mAP@0.5:0.95 \\
    \midrule
    293 & 0.490 & 0.121 & 1.042 
        & 0.909 & 0.181 & 1.122 
        & 0.992 & 0.847  & 0.992   & 0.864      \\
    \bottomrule
  \end{tabular}
\end{table}

\begin{figure}[htbp]
    \centering
    \fbox{%
      \begin{minipage}{\dimexpr\linewidth-2\fboxsep-2\fboxrule\relax}
        \centering
        
        \begin{subfigure}[b]{0.32\linewidth}
            \centering
            \includegraphics[width=\linewidth]{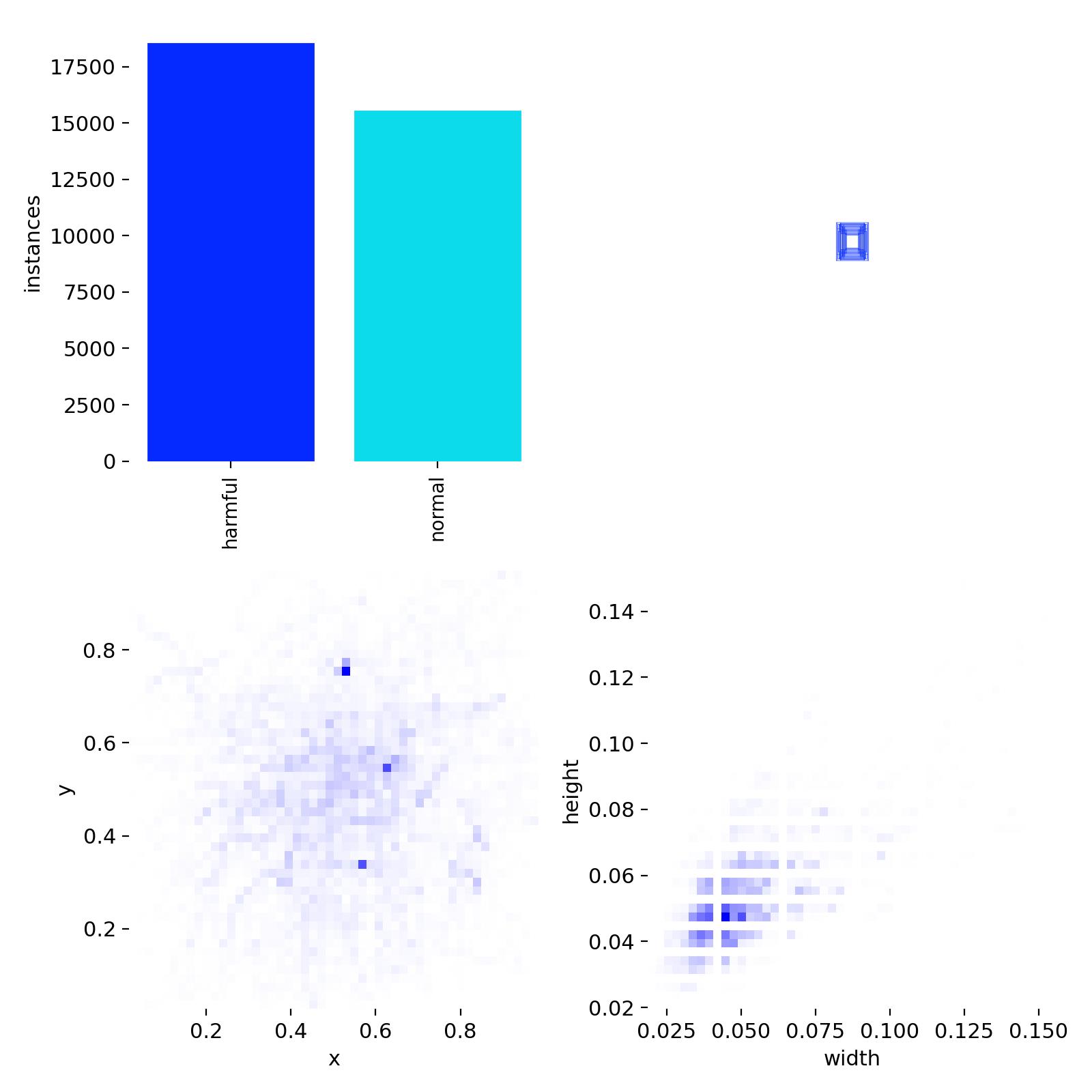}
            \caption{Class distribution \& bbox stats}
            \label{fig:panel:labels}
        \end{subfigure}%
        \hfill%
        \begin{subfigure}[b]{0.32\linewidth}
            \centering
            \includegraphics[width=\linewidth]{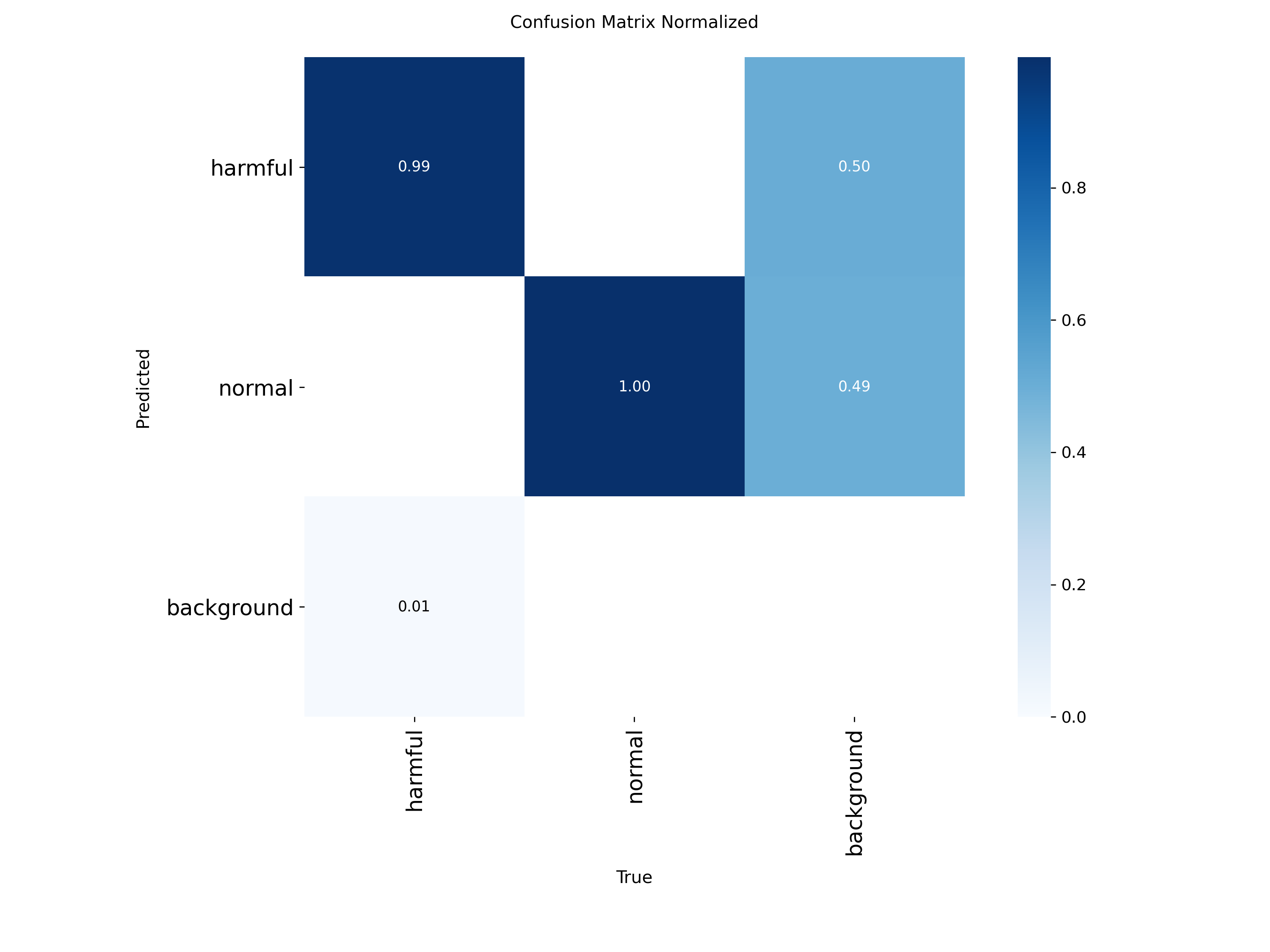}
            \caption{Normalized confusion matrix}
            \label{fig:panel:confusion}
        \end{subfigure}%
        \hfill%
        \begin{subfigure}[b]{0.32\linewidth}
            \centering
            \includegraphics[width=\linewidth]{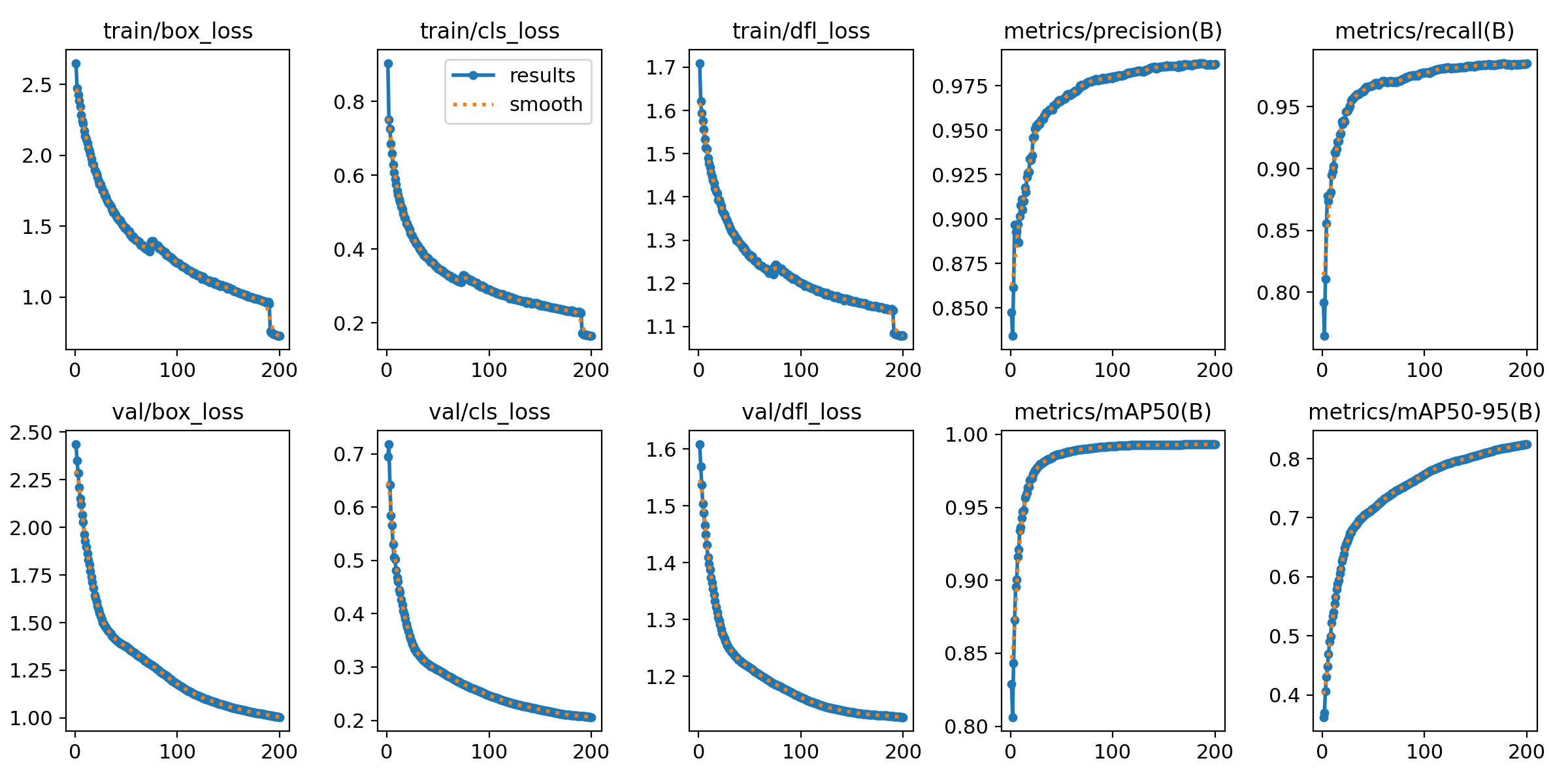}
            \caption{Training \& validation metrics}
            \label{fig:panel:results}
        \end{subfigure}
        
        \vspace{1em}
        
        \begin{subfigure}[b]{0.48\linewidth}
            \centering
            \includegraphics[width=\linewidth]{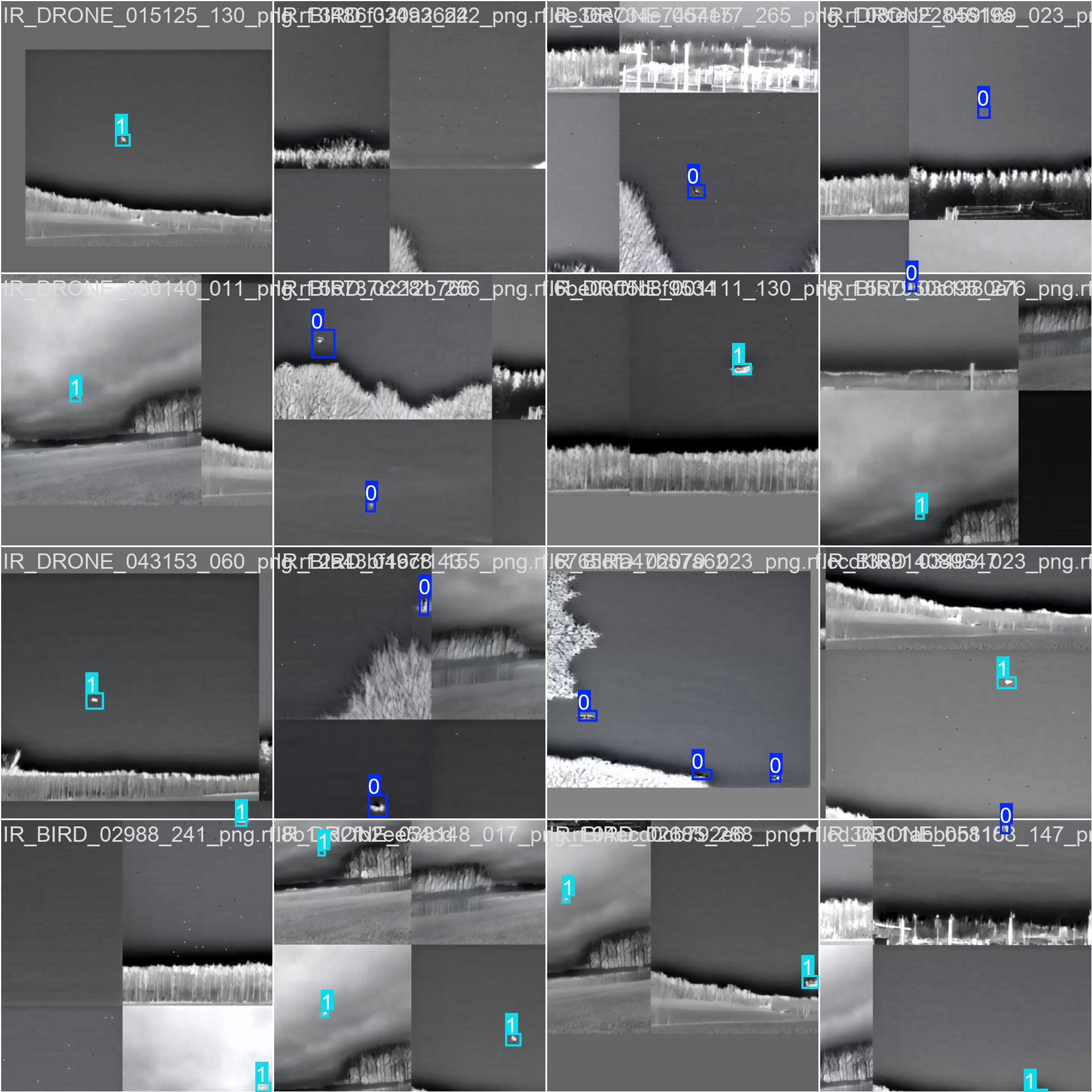}
            \caption{Example training batch}
            \label{fig:panel:train_batch}
        \end{subfigure}%
        \hfill%
        \begin{subfigure}[b]{0.48\linewidth}
            \centering
            \includegraphics[width=\linewidth]{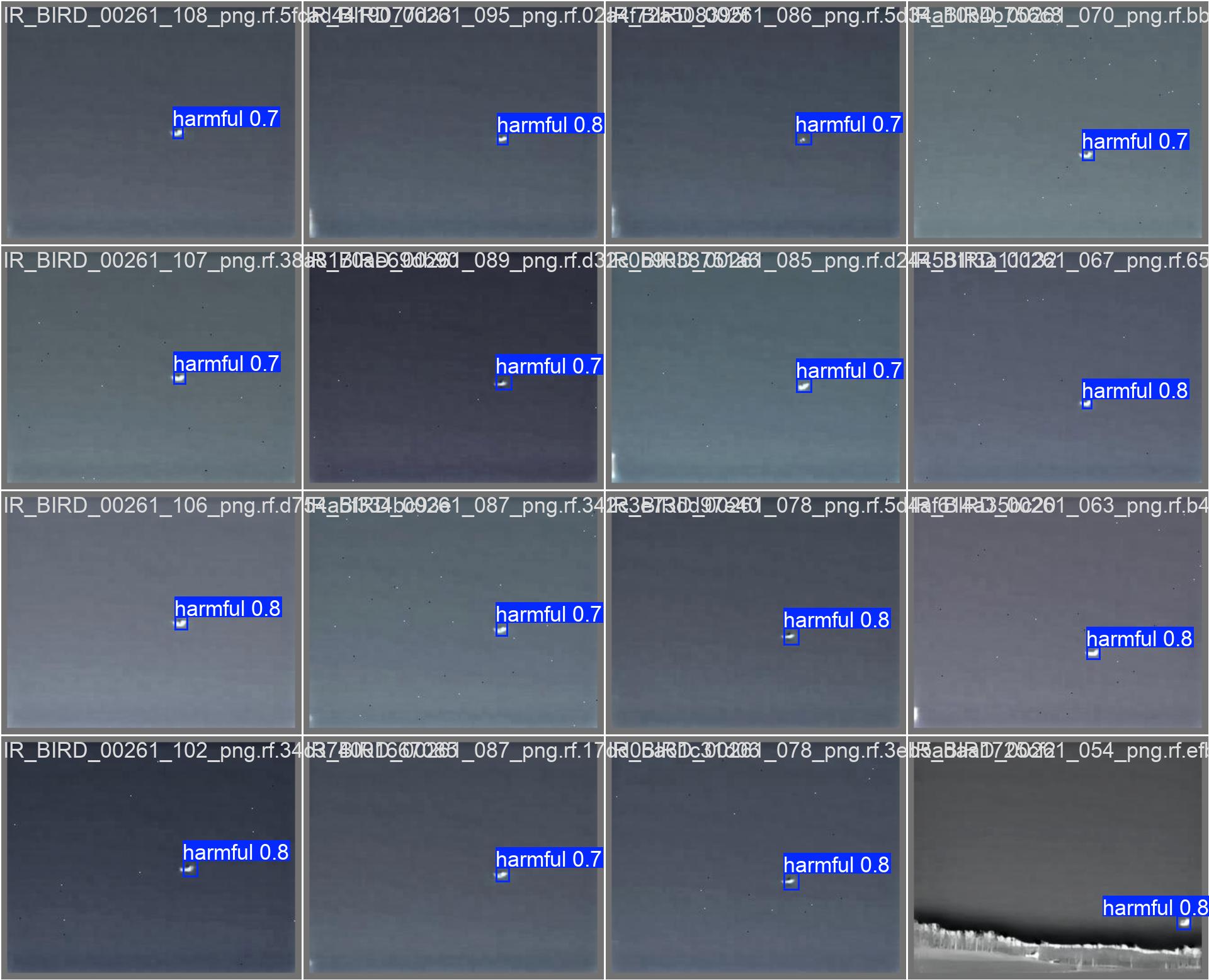}
            \caption{Validation predictions}
            \label{fig:panel:val_pred}
        \end{subfigure}
        
      \end{minipage}%
    }
    \caption[Dataset statistics and model performance overview for training IR Drone Image]{%
      \textbf{Dataset Statistics and Model Performance Overview for training IR Drone Image} 
      (a) Class distribution and bounding-box dimension statistics. 
      (b) Normalized confusion matrix. 
      (c) Evolution of training and validation losses \& metrics. 
      (d) Representative training batch with ground truth. 
      (e) Sample validation detections with confidence scores.%
    }
    \label{fig:dataset_and_results}
\end{figure}

\begin{table}[htbp]
  \centering
  \caption{Payload Classification Performance on IR Dataset}
  \label{tab:ir_payload_performance}
  \begin{tabular}{lcc cc ccc}
    \toprule
    & \multicolumn{2}{c}{\textbf{Raw Counts}} 
    & \multicolumn{2}{c}{\textbf{Normalized}} 
    & \multicolumn{3}{c}{\textbf{Metrics}} \\
    \cmidrule(r){2-3} \cmidrule(r){4-5} \cmidrule(l){6-8}
    \textbf{Dataset} & Harmful & Normal & Harmful & Normal & Prec. & Rec. & F$_1$ \\
    \midrule
    IR & 1\,992 & 2\,033 & 0.99 & 0.99 & 0.99 & 0.99 & 0.99 \\
    \bottomrule
  \end{tabular}
\end{table}

\begin{figure}[htbp]
    \centering
    \fbox{%
      \begin{minipage}{\dimexpr\linewidth-2\fboxsep-2\fboxrule\relax}
        \centering
        
        \begin{subfigure}[b]{0.32\linewidth}
            \centering
            \includegraphics[width=\linewidth]{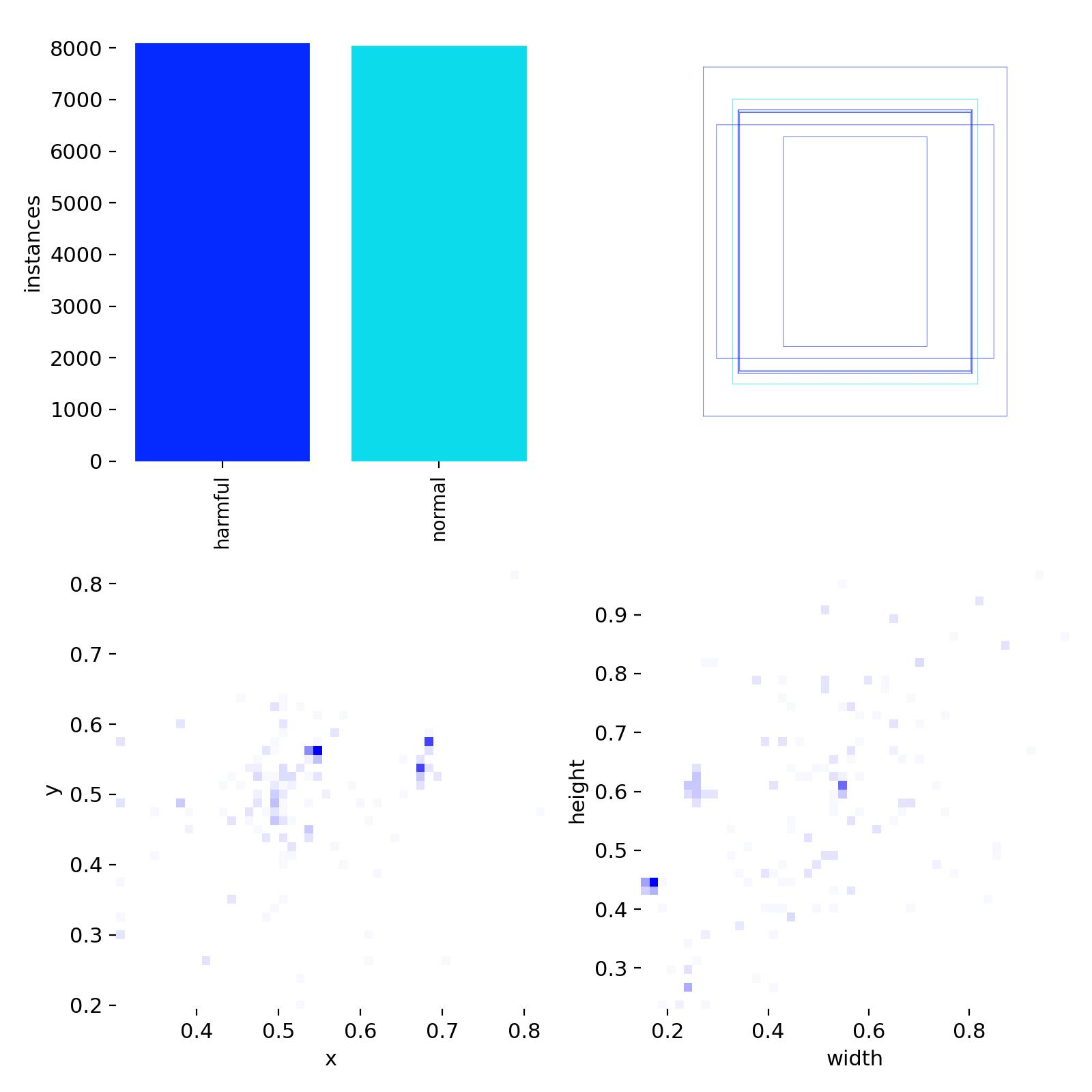}
            \caption{Class distribution \& bbox stats}
            \label{fig:panel:labels}
        \end{subfigure}%
        \hfill%
        \begin{subfigure}[b]{0.32\linewidth}
            \centering
            \includegraphics[width=\linewidth]{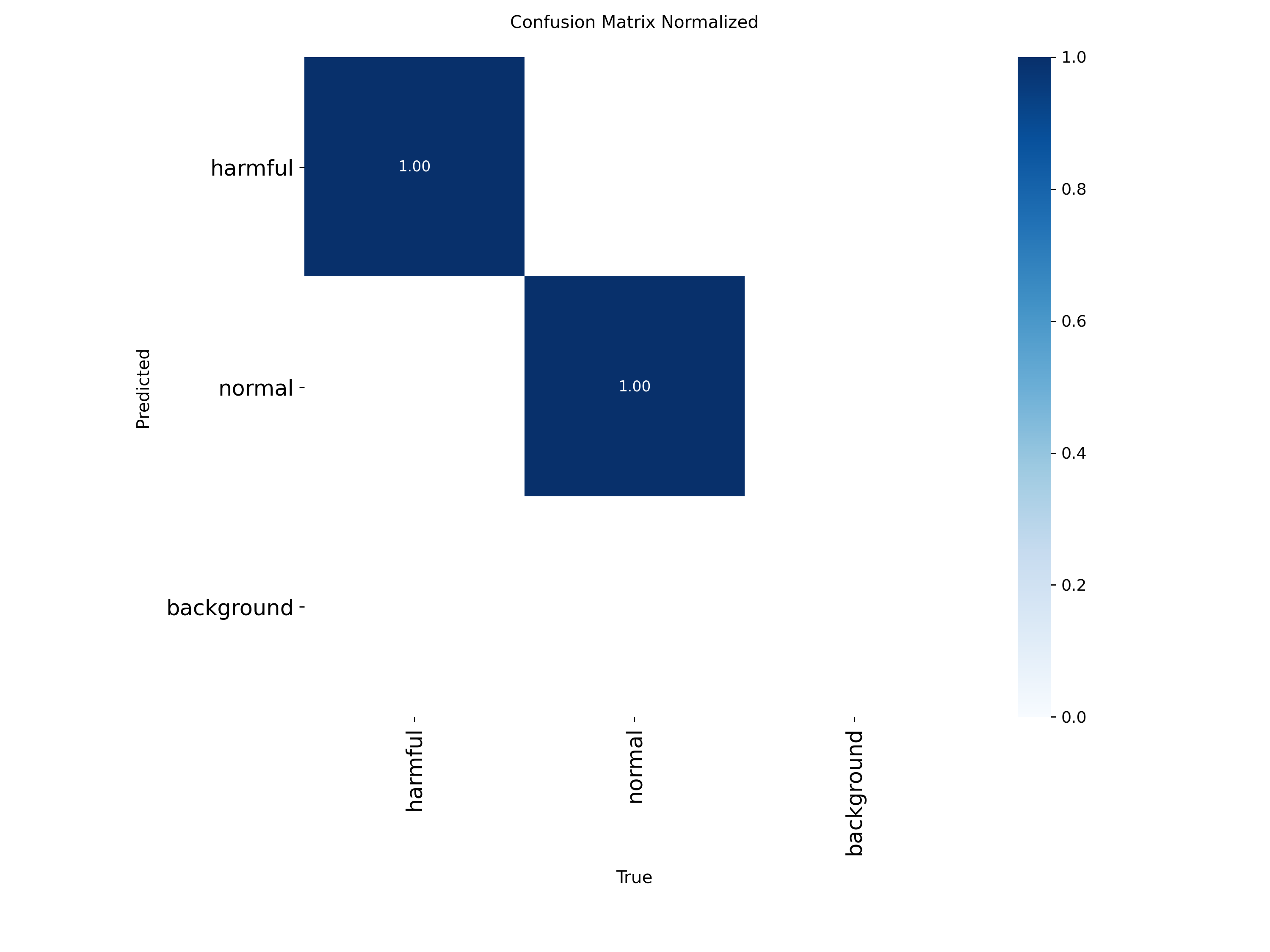}
            \caption{Normalized confusion matrix}
            \label{fig:panel:confusion}
        \end{subfigure}%
        \hfill%
        \begin{subfigure}[b]{0.32\linewidth}
            \centering
            \includegraphics[width=\linewidth]{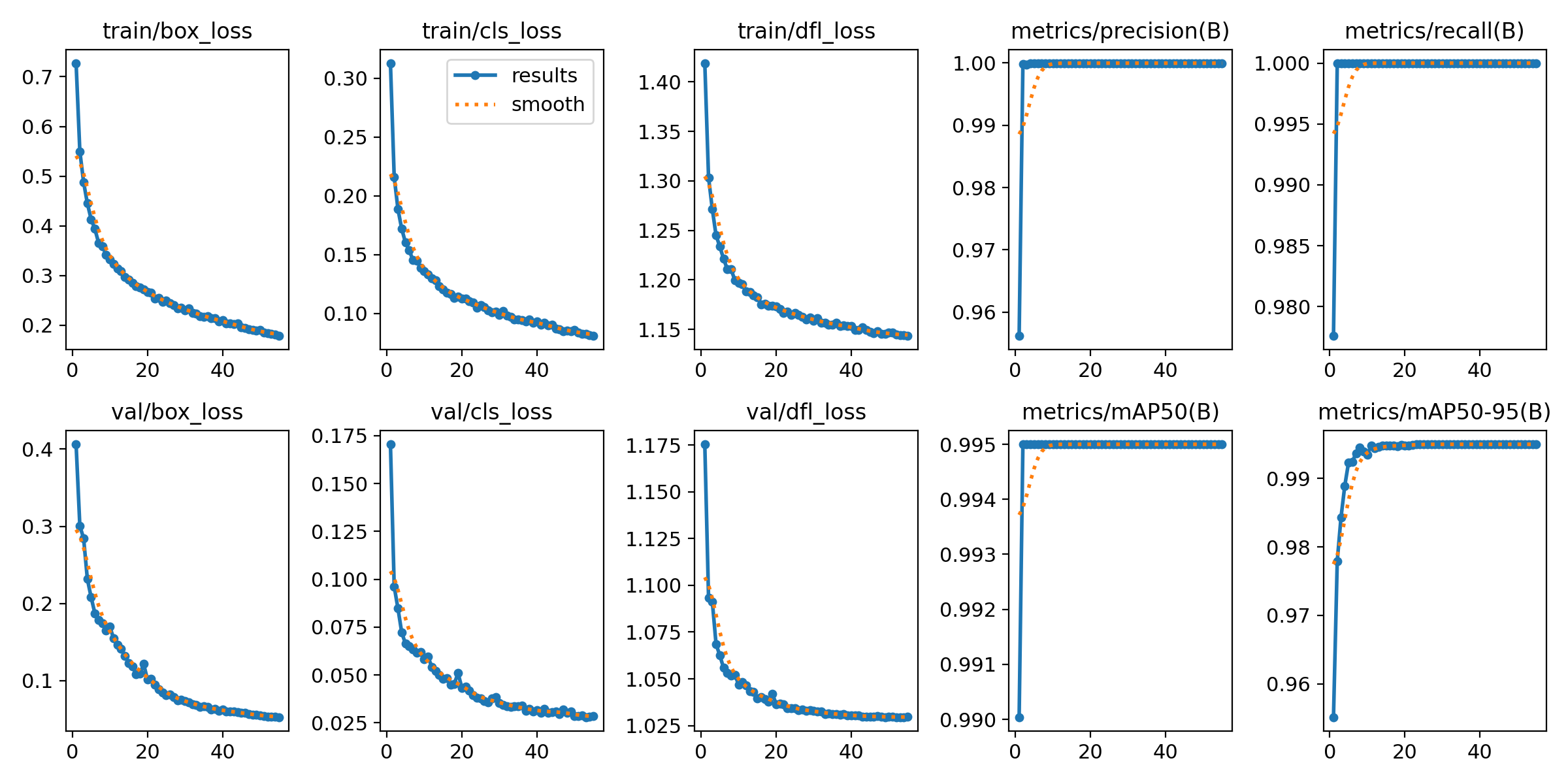}
            \caption{Training \& validation metrics}
            \label{fig:panel:results}
        \end{subfigure}
        
        \vspace{1em}
        
        \begin{subfigure}[b]{0.48\linewidth}
            \centering
            \includegraphics[width=\linewidth]{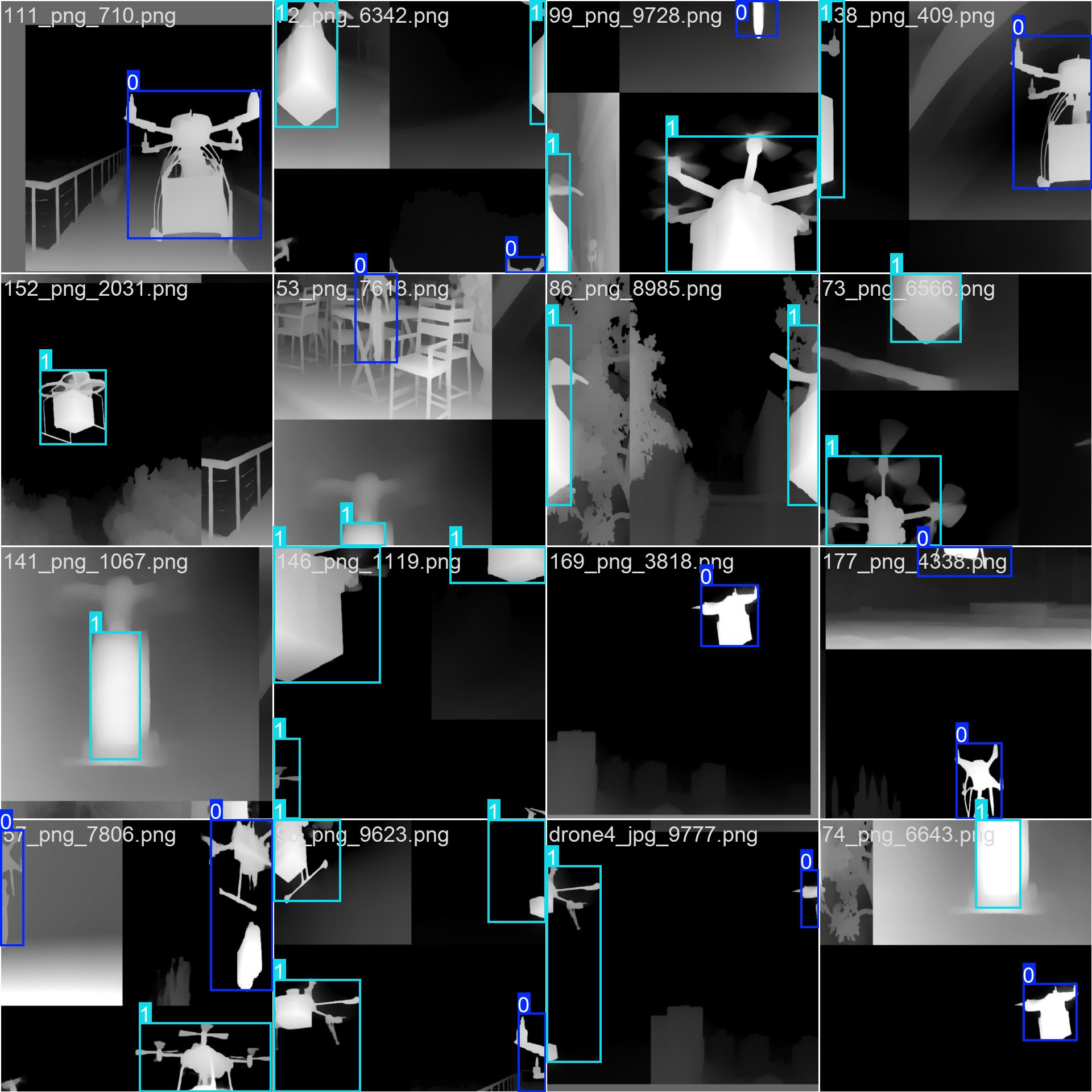}
            \caption{Example training batch}
            \label{fig:panel:train_batch}
        \end{subfigure}%
        \hfill%
        \begin{subfigure}[b]{0.48\linewidth}
            \centering
            \includegraphics[width=\linewidth]{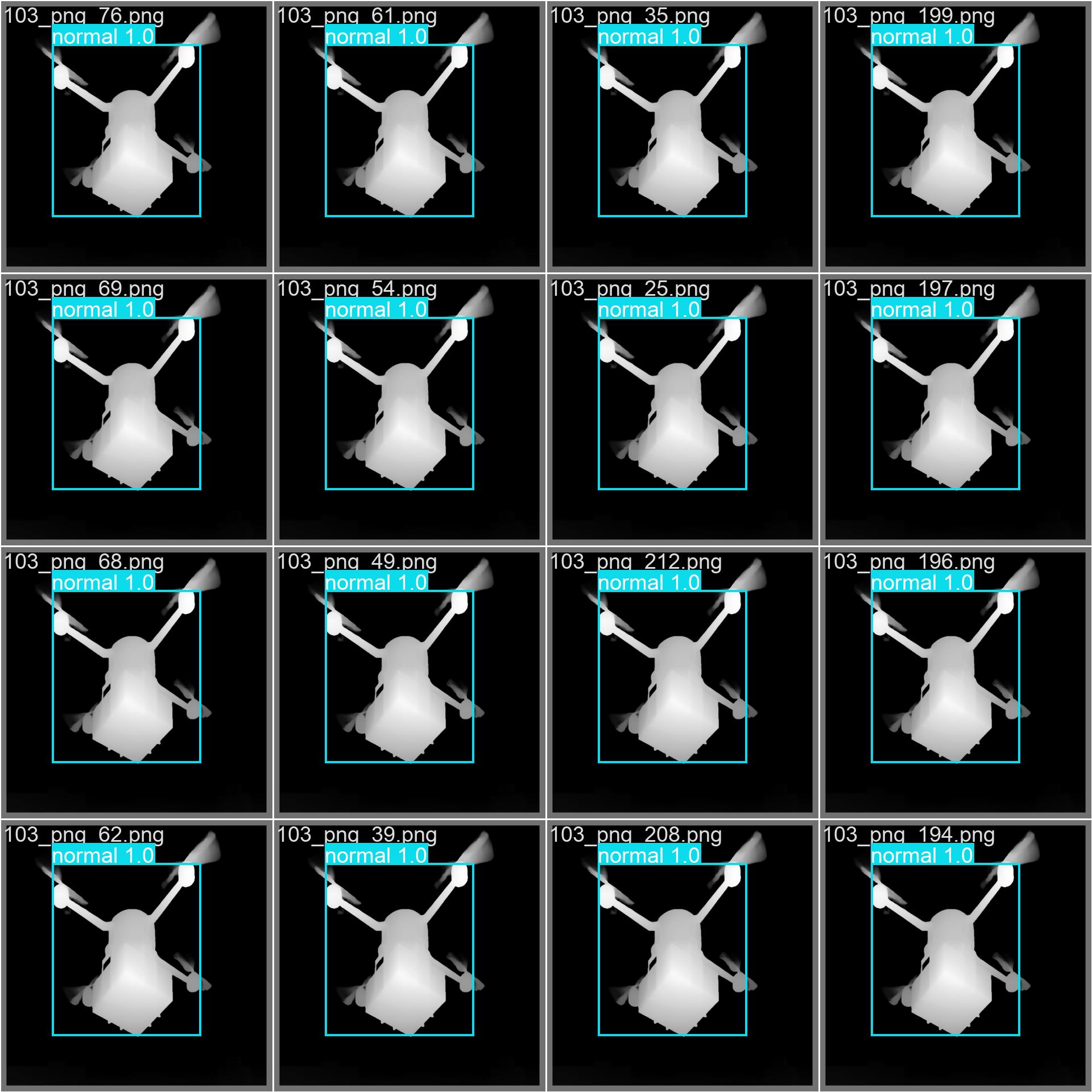}
            \caption{Validation predictions}
            \label{fig:panel:val_pred}
        \end{subfigure}
        
      \end{minipage}%
    }
    \caption[Dataset statistics and model performance overview for training payload IR  Image]{%
      \textbf{Dataset Statistics and Model Performance Overview for training payload IR  Image} 
      (a) Class distribution and bounding-box dimension statistics. 
      (b) Normalized confusion matrix. 
      (c) Evolution of training and validation losses \& metrics. 
      (d) Representative training batch with ground truth. 
      (e) Sample validation detections with confidence scores.%
    }
    \label{fig:dataset_and_results}
\end{figure}

\paragraph{Quantitative Results.}
Table~\ref{tab:midfusion_results} presents the performance outcomes of the tested fusion approaches.

\begin{table}[htbp]
\centering
\caption{Performance of Fusion Strategies on Drone and Bird Detection}
\label{tab:midfusion_results}
\begin{tabular}{|l|c|c|}
\hline
\textbf{Model} & \textbf{Epochs} & \textbf{mAP@0.5:0.95} \\
\hline
Mid-Fusion (Basic) & 100 & 0.81 \\
Mid-to-Late Fusion & 100 & 0.78 \\
Mid-Fusion + Transformer & 150 & \textbf{0.84} \\
\hline
\end{tabular}
\end{table}

\paragraph{Late Fusion with Decision Routing.}
Subsequently, we explored a late fusion strategy based on a modular decision-layer routing mechanism, as illustrated in Fig.~2. Under this setup, RGB and IR streams were independently processed by their respective YOLO11n backbones. During inference, each modality produced independent bounding box predictions and associated confidence scores. These outputs were dynamically selected by a decision layer that employed heuristic rules based on object activation and modality confidence, effectively choosing between RGB, IR, or both depending on input reliability and environmental conditions.

\begin{table}[htbp]
\centering
\caption{Validation Results for Independently Trained RGB and IR YOLO11n Models }
\label{tab:individual_modalities}
\begin{tabular}{|l|c|c|c|c|c|c|}
\hline
\textbf{Modality} & \textbf{mAP@0.5:0.95} & \textbf{mAP@0.5} & \textbf{mAP@0.75} & \textbf{Precision} & \textbf{Recall} & \textbf{F1-Score} \\
\hline
RGB (YOLO11n) & 0.807 & 0.989 & 0.897 & 0.981 & 0.972 & 0.977 \\
IR (YOLO11n)  & 0.81 & 0.988 & 0.896 & 0.978 & 0.97 & 0.976 \\
\hline
\end{tabular}
\end{table}

\paragraph{Fusion via Decision Layer.}
In the late fusion mechanism, we employed weighted confidence selection and Non-Maximum Suppression (NMS) to efficiently combine bounding box predictions from both modalities. The decision layer dynamically favored the modality providing higher confidence or better contextual conditions—RGB was preferred for drones under IR-unfavorable conditions, while IR often detected occluded or poorly illuminated birds more reliably. This fusion method notably reduced redundant detections without the need for retraining or parameter integration. Despite individually high-performing models, the fusion achieved a slightly lower compared to transformer-based mid-level fusion.

\paragraph{Analysis.}
Our results indicated that mid-level fusion approaches—particularly transformer-enhanced mid-fusion—demonstrated superior performance due to early and sophisticated integration of modality-specific features through attention mechanisms. While the late fusion via decision routing provided adaptability to varying conditions, the absence of jointly optimized training limited its overall effectiveness. These insights highlight transformer-based mid-level fusion as a highly promising approach for multimodal object detection. Future investigations may include dynamic fusion weighting and advanced spatiotemporal attention strategies to further enhance detection performance.

\subsection{Payload Detection}

\begin{figure}[h]
    \centering
    \includegraphics[width=0.75\linewidth]{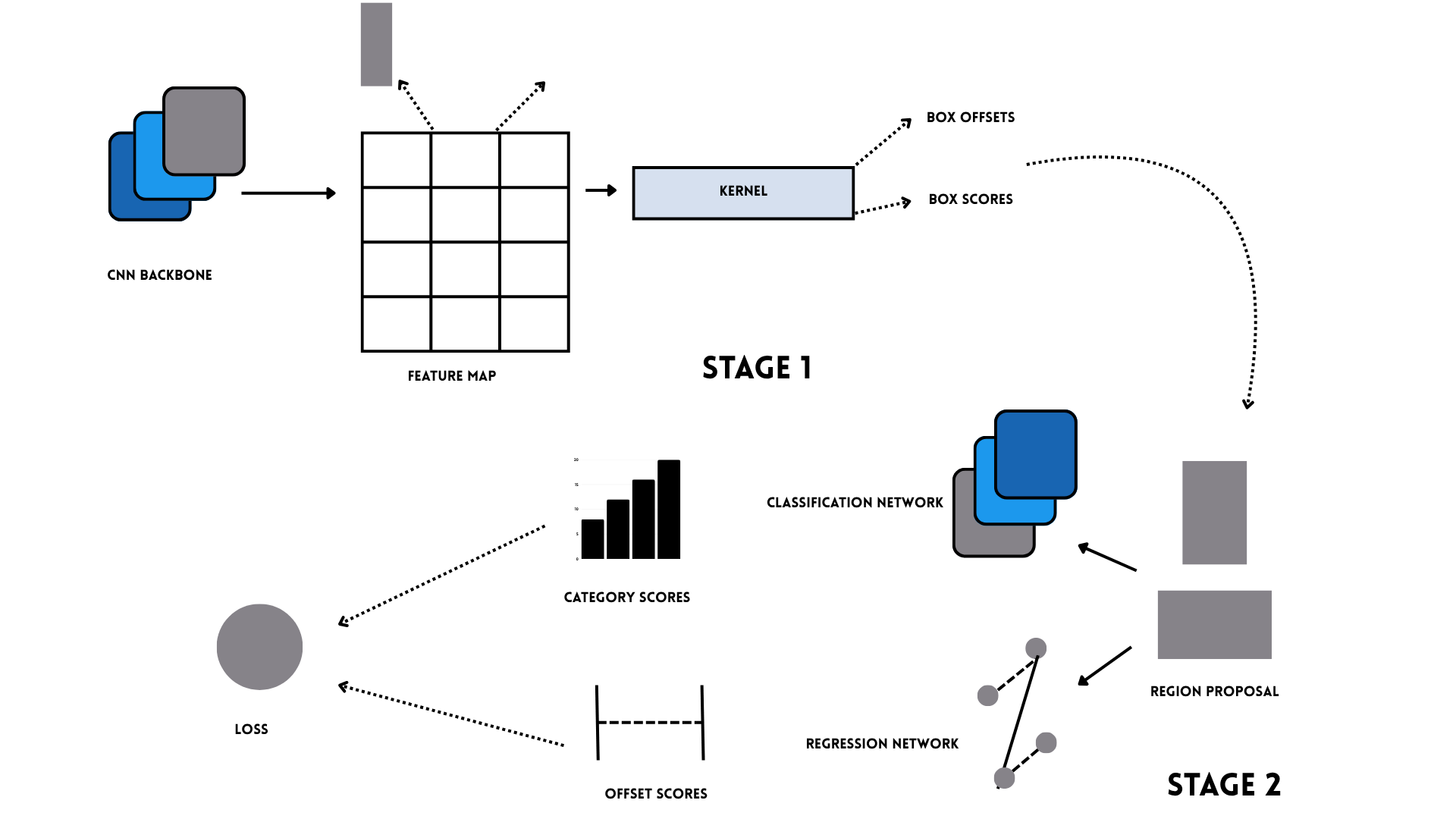}
    \caption{faster R-CNN Overview}
    \label{fig:single_flow}
\end{figure}

For payload classification under IR and RGB modalities, we firstly employed the Faster R-CNN architecture built upon a ResNet-50 backbone with Feature Pyramid Networks (FPN), as implemented in the torchvision detection module. The model was pretrained on the COCO dataset and fine-tuned for our 3-class problem: background, harmful payloads, and normal payloads. The classification head was replaced using the \texttt{FastRCNNPredictor} to match our custom label set. Training was conducted separately on RGB and IR image sets, using YOLO-format bounding box annotations converted to Pascal VOC format for compatibility. A small batch size was used due to GPU memory constraints, and Gaussian noise was applied as augmentation to improve robustness.

Table~\ref{tab:fasterrcnn_val_results} summarizes the validation performance of the trained models on both modalities. 

\begin{table}[htbp]
\centering
\caption{Validation Results of Faster R-CNN Models for Payload Detection}
\label{tab:fasterrcnn_val_results}
\begin{tabular}{|l|c|c|c|c|c|}
\hline
\textbf{Modality} & \textbf{mAP@0.5:0.95} & \textbf{mAP@0.5} & \textbf{Precision} & \textbf{Recall} & \textbf{F1-Score} \\
\hline
RGB (Faster R-CNN) & 0.997 & 0.994 & 0.930 & 0.915 & 0.969 \\
IR (Faster R-CNN)  & 0.950 & 0.964 & 0.949 & 0.938 & 0.943 \\
\hline
\end{tabular}
\end{table}
Following this, we leveraged a lightweight YOLOv11n dual-backbone architecture with decision-layer routing for real-time inference. The architecture, depicted in Fig. 3, supports three modes: RGB-only, IR-only, and RGB-IR. A placeholder channel injection mechanism ensures consistent tensor shapes across modalities. Each backbone is trained separately, and during inference, outputs are passed through activation-based Non-Maximum Suppression (NMS) and a decision-layer module to yield final predictions.

This architecture led to substantial improvements in inference accuracy across all modalities. Table~\ref{tab:yolov11_val_results} presents the validation scores.

\begin{table}[htbp]
\centering
\caption{YOLOv11n Validation Results Using Decision-Layer Late Fusion}
\label{tab:yolov11_val_results}
\begin{tabular}{|l|c|c|c|c|c|}
\hline
\textbf{Modality} & \textbf{mAP@0.5:0.95} & \textbf{mAP@0.5} & \textbf{Precision} & \textbf{Recall} & \textbf{F1-Score} \\
\hline
RGB (YOLOv11n) & 0.989 & 0.995 & 0.997 & 0.998 & 0.999 \\
IR (YOLOv11n)  & 0.990 & 0.995 & 0.969 & 0.988 & 0.993 \\

\hline
\end{tabular}
\end{table}

\section{Conclusion}

In this work, we presented \textit{SpectraSentinel}, a lightweight, real-time dual-stream system for drone detection, tracking, and payload identification using RGB and infrared (IR) modalities. Our approach integrates modality-specific YOLOv11n detectors, efficient tracking via DeepSORT, and a hybrid direction estimation pipeline to address the challenges of small aerial object surveillance in complex environments. By processing RGB and IR streams independently and fusing their outputs at the decision level, we achieve robustness to modality-specific limitations—enhancing detection in low-visibility scenarios and improving payload recognition through complementary cues.

We further explored various fusion strategies—ranging from mid-level transformer-based fusion to modular late fusion using decision layers—and demonstrated that mid-level fusion yields superior accuracy, while late fusion provides architectural flexibility and operational adaptability. Extensive experimental results on the VIP Cup 2025 dataset validate the effectiveness of our models, showing strong performance across detection, tracking, and payload classification tasks, even under adverse conditions.

Our system is designed with real-time constraints in mind, making it deployable on edge devices for practical surveillance scenarios. Looking ahead, future work may explore joint end-to-end training of multi-modal detectors, advanced spatiotemporal reasoning for behavior prediction, and integration with broader UAV traffic management systems. The promising results of \textit{SpectraSentinel} underline the potential of dual-modality vision systems for intelligent aerial threat monitoring and situational awareness.


\bibliographystyle{ieeetr}

\bibliographystyle{IEEEtran}

\end{document}